\documentclass[accepted]{uai2022} 

\usepackage[american]{babel}

\usepackage{natbib}[compress] 
\usepackage{mathtools} 
\usepackage{booktabs} 
\usepackage{tikz} 
\usepackage{hyperref}       
\usepackage{url}            
\usepackage{amsfonts}       
\usepackage{amssymb}
\usepackage{amsmath}
\usepackage{nicefrac}       
\usepackage{microtype}      
\usepackage{xcolor}         
\usepackage{graphicx}
\usepackage{xspace}
\usepackage{bm}
\usepackage{soul}
\usepackage{caption}
\usepackage{subcaption}
\usepackage{wrapfig}
\usepackage{placeins}
\usepackage{enumitem}
\usepackage{footnote}
\usepackage{amsthm}
\usepackage{xr}

\makesavenoteenv{tabular}
\makesavenoteenv{table}
\theoremstyle{plain}
\newtheorem{theorem}{Theorem}[section]
\newtheorem{manualtheoreminner}{Theorem}
\newenvironment{manualtheorem}[1]{%
  \renewcommand\themanualtheoreminner{#1}%
  \manualtheoreminner
}{\endmanualtheoreminner}
\newtheorem{manuallemmainner}{Lemma}
\newenvironment{manuallemma}[1]{%
  \renewcommand\themanuallemmainner{#1}%
  \manuallemmainner
}{\endmanuallemmainner}

\newtheorem{lemma}{Lemma}[section]

\theoremstyle{definition}
\newtheorem{definition}{Definition}[section]

\theoremstyle{remark}

\usepackage[linesnumbered,ruled,vlined]{algorithm2e}
\newtheorem{innercustomgeneric}{\customgenericname}
\providecommand{\customgenericname}{}
\newcommand{\newcustomtheorem}[2]{%
  \newenvironment{#1}[1]
  {%
   \renewcommand\customgenericname{#2}%
   \renewcommand\theinnercustomgeneric{##1}%
   \innercustomgeneric
  }
  {\endinnercustomgeneric}
}

\newcustomtheorem{customthm}{Theorem}
\newcustomtheorem{customlemma}{Lemma}

\DeclareMathOperator*{\argmax}{arg\,max}

\makeatletter
\newcommand*{\addFileDependency}[1]{
  \typeout{(#1)}
  \@addtofilelist{#1}
  \IfFileExists{#1}{}{\typeout{No file #1.}}
}
\makeatother

\newcommand*{\myexternaldocument}[1]{%
    \externaldocument{#1}%
    \addFileDependency{#1.tex}%
    \addFileDependency{#1.aux}%
}
\myexternaldocument{daulton_446-supp-input}

\newcommand{\ALG}{MORBO\xspace}
\newcommand{\HV}{\textsc{HV}}
\newcommand{\HVI}{\textsc{HVI}}
\newcommand{\HVC}{\textsc{HVC}}

\title{Multi-Objective Bayesian Optimization over High-Dimensional Search Spaces}

\author[*,1,2]{\href{mailto:<sdaulton@fb.com>?Subject=Your UAI 2022 paper}{Samuel Daulton}{}}
\author[*,2]{David Eriksson}
\author[2]{Maximillian Balandat}
\author[2]{Eytan Bakshy}
\affil[*]{%
    Equal contribution
}
\affil[1]{%
    University of Oxford\\
    Oxford, UK
}
\affil[2]{%
    Meta\\
    Menlo Park, USA
}

\begin{document}
\maketitle

\begin{abstract}
    Many real world scientific and industrial applications require optimizing multiple competing black-box objectives.
    When the objectives are expensive-to-evaluate, multi-objective Bayesian optimization (BO) is a popular approach because of its high sample efficiency.
    However, even with recent methodological advances, most existing multi-objective BO methods perform poorly on search spaces with more than a few dozen parameters and rely on global surrogate models that scale cubically with the number of observations.
    In this work we propose \ALG, a scalable method for multi-objective BO over high-dimensional search spaces.
    \ALG identifies diverse globally optimal solutions by performing BO in multiple local regions of the design space in parallel using a coordinated strategy.
    We show that \ALG significantly advances the state-of-the-art in sample efficiency for several high-dimensional synthetic problems and real world applications, including an optical display design problem and a vehicle design problem with $146$ and $222$ parameters, respectively.
    On these problems, where existing BO algorithms fail to scale and perform well, \ALG provides practitioners with order-of-magnitude improvements in sample efficiency over the current approach.
\end{abstract}
\section{Introduction}
The challenge of identifying optimal trade-offs between multiple complex objective functions is pervasive in many fields, including machine learning~\citep{sener2018mtmoo}, science~\citep{gopakumar2018moomaterial}, and engineering~\citep{marler2004survey,mathern2021}.
For instance, Mazda recently proposed a vehicle design problem in which the goal is to optimize the widths of $222$ structural parts in order to minimize the total weight of three different vehicles while simultaneously maximizing the number of common gauge parts~\citep{kohira2018proposal}.
Additionally, this problem has $54$ black-box constraints that enforce important performance requirements such as collision safety.
Evaluating a design requires either crash-testing a physical prototype or running computationally demanding simulations.
In fact, the original problem was solved on what at the time was the world's fastest supercomputer and took around {$3$,$000$} CPU years to compute~\citep{oyama2017mazda}.
Another example is designing optical components for AR/VR applications, which requires optimizing complex geometries described by hundreds of parameters in order to identify designs that yield optimal trade-offs between image quality and efficiency of the optical device.
Evaluating a design involves either fabricating and measuring prototypes or running computationally intensive simulations.
For such problems, sample-efficient optimization is paramount.

Bayesian optimization (BO) has emerged as an effective, general, and sample-efficient approach for ``black-box'' optimization~\citep{jones98} and is highly effective for machine learning hyperparameter tuning~\citep{turner2021bayesian}.
However, in its basic form, BO is subject to important limitations.
In particular, (i) successful applications typically consider low-dimensional search spaces, usually with less than $20$ tunable parameters~\citep{frazier2018tutorial}, (ii) inference with the typical Gaussian Process (GP) surrogate models incurs cubic time complexity with respect to the number of data points, which prevents usage in the large-sample regime that is often necessary for high-dimensional problems, and (iii) most methods focus on single objective unconstrained problems.
As a result, BO cannot easily be applied to either of the aforementioned Mazda vehicle design or the AR/VR optical design problems. Moreover, high dimensional multi-objective problems requiring sample-efficient optimization are prevalent in many real-world settings such as groundwater remediation \citep{akhtar2015}, cell network configuration \citep{dreifuerst2021optimizing}, and water resource management \citep{bai17}.
The state-of-the-art approach for this class of problems is NSGA-II~\citep{deb02nsgaii}, a popular evolutionary strategy, but with poor sample-efficiency, which hinders the progress of the scientists running these experiments.

\begin{figure*}[!ht]
    \centering
    \includegraphics[width=0.96\textwidth]{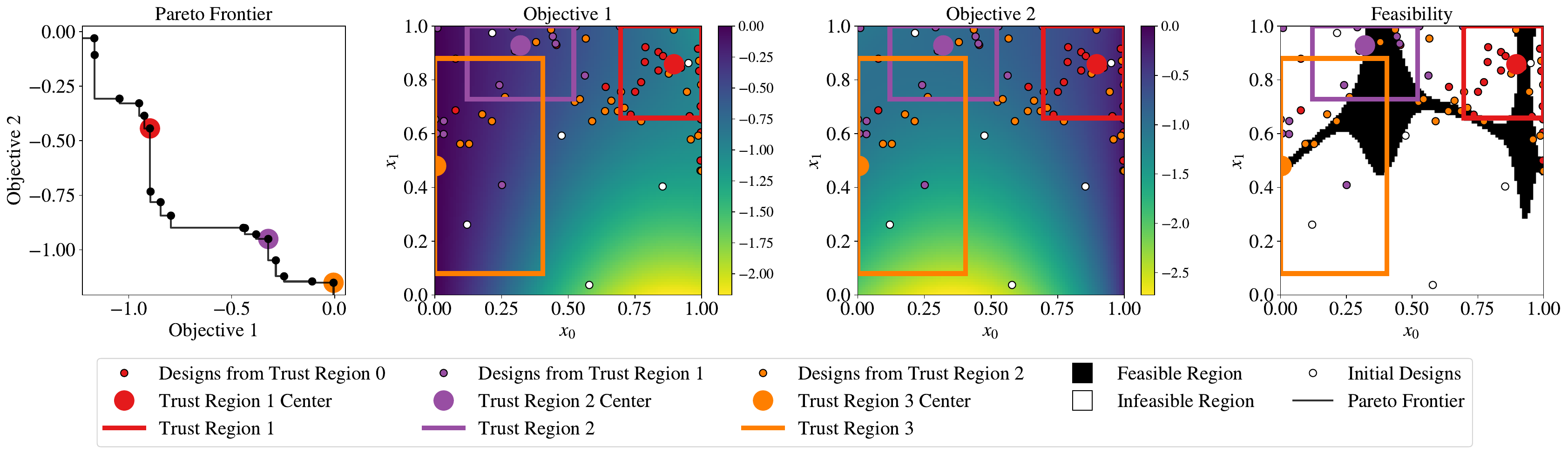}
    \caption{An illustration of \ALG on: 2-objective benchmark problem with 2 parameters and 2 constraints called MW7 \citep{mw_test_problems} with $3$ TRs. The left-most plot illustrates how \ALG's center selection technique centers the TRs at Pareto optimal points across different parts of the Pareto frontier. This encourages \ALG{} to explore diverse parts of the Pareto Frontier, which is important to identifying the multiple disconnected regions on this MW7 problem. The three right-most plots illustrate the TRs over the design space along with contours of, respectively, the 2 objective metrics and the feasibility metric indicating whether all black-box constraints are satisfied. Note that the TRs overlap with one another and contain data points that were collected by other TRs. Hence, sharing observations collected by different TRs provides local models with more  observations than if each local model were only fitted to data collected by its corresponding TR.}
    \label{fig:conceptual}
\end{figure*}

In this paper, we close this gap by making BO applicable to challenging high-dimensional multi-objective problems.
To do so, we propose an algorithm called \ALG (``Multi-Objective Regionalized Bayesian Optimization'') that optimizes diverse parts of the global Pareto frontier in parallel using a coordinated set of local trust regions (TRs).
As shown in Figure~\ref{fig:conceptual}~(left), TRs are located at different solutions with diverse trade-offs between objectives.
\ALG{} performs local BO in each TR to mitigate over-exploration, a phenomenon that plagues many algorithms in high-dimensional settings~\citep{eriksson2021scalable}.
To enable scaling to large evaluation budgets, \ALG{} leverages \emph{local} GP surrogate models of the objective function, which reduces the time complexity for GP inference from $O(n^3)$, where $n$ is the number of data points, to $O(n_{\mathcal T}^3)$, where $n_{\mathcal T} \ll n$ is the number of local data points for a TR~$\mathcal T$.
To facilitate efficient and collaborative global optimization, \ALG{} \emph{passes information} between TRs in the following two ways: (1) Observations collected by one TR are shared with the others---which is particularly important when the TRs overlap as shown in Figure~\ref{fig:conceptual}, (2) \ALG{} selects a batch of candidates by leveraging the TRs to collaboratively maximize a global utility.
To ensure efficient global optimization, \ALG terminates under-performing TRs and allocates new TRs according to a global policy with a theoretical performance guarantee---a property that sets \ALG apart from most existing methods.

The significance of \ALG is that it is the \emph{first} multi-objective BO method that scales to hundreds of tunable parameters and thousands of evaluations, a setting where practitioners have previously had to fall back on alternative methods with much lower sample-efficiency, such as NSGA-II. Our comprehensive evaluation demonstrates that \ALG yields \emph{order-of-magnitude} savings in terms of time and resources compared to state-of-the-art methods on challenging high-dimensional multi-objective problems.

\section{Background}
\label{sec:background}
\subsection{Preliminaries}

\subsubsection{Multi-Objective Optimization}
In multi-objective optimization (MOO), the goal is to maximize (without loss of generality) a vector-valued objective function $\bm f(\bm x) = [f^{(1)}(\bm x), ..., f^{(M)}(\bm x)] \in \mathbb{R}^M$, where $M\geq2$ while satisfying black-box constraints $\bm g(\bm x) \geq \bm 0 \in \mathbb R^V$ where $V\geq 0$, $\bm x \in \mathcal X \subset \mathbb{R}^d$, and $\mathcal X $ is a compact set.
Usually, there is no single solution $\bm x^*$ that simultaneously maximizes all $M$ objectives and satisfies all $V$ constraints.
Hence, objective vectors are compared using Pareto domination.
\begin{definition}
An objective vector $\bm f(\bm x)$ \emph{Pareto-dominates} 
$\bm f(\bm x')$, denoted as $\bm f(\bm x)\succ \bm f(\bm x')$,
if $f^{(m)}(\bm x) \geq f^{(m)}(\bm x')$ for all $m=1, ..., M$ and there exists at least one $m \in \{1, \dotsc, M\}$ such that $f^{(m)}(\bm x) > f^{(m)}(\bm x')$.
\end{definition}
\begin{definition}
The \emph{Pareto frontier} (PF) is the set of optimal trade-offs $\mathcal P(X)$ over a set of designs $X \subseteq \mathcal X$:
$$\mathcal P(X) = \{\bm f(\bm x) : \bm x \in X, \nexists ~\bm x' \in X ~s.t.~ \bm f(\bm x') \succ \bm f(\bm x)\}$$
Under black-box constraints, the \emph{feasible Pareto frontier} is defined as $\mathcal P_\text{feas}(X)$ = $\mathcal P(\{\bm x \in X : \bm g(\bm x) \geq \bm 0\})$.
\end{definition}
The goal of a MOO algorithm is to identify an approximate PF $\mathcal P(X_n)$ of the true PF $\mathcal P(\mathcal X)$ within a pre-specified budget of $|X_n| = n$ function evaluations. The quality of a PF is often evaluated using the hypervolume (\HV{}) indicator.
\begin{definition}
The \emph{hypervolume indicator}, $\HV{}(\mathcal P(X) | \bm r)$ is the $M$-dimensional Lebesgue measure $\lambda_M$ of the region dominated by $\mathcal P(X)$ and bounded from below by a reference point $\bm r \in \mathbb R^M$.
\end{definition}

The reference point is typically provided by the practitioner based on domain knowledge~\citep{yang2019}.
MOO problems are often addressed using evolutionary algorithms (EA) such as NSGA-II~\citep{deb02nsgaii}.
However, EAs generally suffer from high sample-complexity, rendering them inapplicable under small evaluation budgets.
\subsubsection{Bayesian Optimization}
When high sample-efficiency is required, Bayesian optimization (BO) is a popular approach \citep{frazier2018tutorial}.
BO relies on a probabilistic surrogate model and an acquisition function that uses the surrogate model to provide the utility of evaluating a set of design points on the black-box function. The acquisition function is responsible for balancing exploration and exploitation.
In the multi-objective setting, a common approach is to optimize random scalarizations of the objectives~\citep{parego, paria2020flexible} using a single-objective acquisition function.
A more principled approach is to directly optimize the Pareto frontier by selecting candidates with maximum hypervolume improvement either in expectation under the GP posterior~\citep{emmerich2006} or using Thompson sampling (TS) \citep{tsemo}.
\subsection{Related Work}
\subsubsection{Multi-objective Bayesian optimization}
There have been many recent contributions to multi-objective BO, e.g.,~\citet{konakovic2020diversity, daulton2020ehvi, daulton2022robust,tsemo}), but very few methods consider the high-dimensional setting and with large evaluation budgets. All of these methods described below rely on global GP models.
As a result, these methods have mostly been evaluated on low-dimensional problems, typically $d \ll 10$ \citep{konakovic2020diversity, tsemo}.
In the multi-objective BO literature, the largest search space we have found consists of $27$ parameters~\citep{paria2020flexible}.
Nevertheless, for completeness we review multi-objective BO methods that support generating large batches of designs.
DGEMO \citep{konakovic2020diversity} uses a hypervolume-based objective with heuristics to encourage diversity while exploring the PF.

Parallel expected hypervolume improvement ($q$EHVI)~\citep{daulton2020ehvi} has strong empirical performance, but its computational complexity scales exponentially with the batch size.
$q$NEHVI~\citep{daulton2021nehvi} improves scalability with respect to the batch size, but like DGEMO and $q$EHVI, $q$NEHVI has only been evaluated on low-dimensional search spaces.
TSEMO~\citep{tsemo} optimizes approximate GP function samples using NSGA-II and uses a hypervolume-based objective for selecting a batch of points from the NSGA-II population.
ParEGO~\citep{parego} and TS-TCH~\citep{paria2020flexible} use random Chebyshev scalarizations with parallel expected improvement~\citep{jones98} and Thompson sampling---where a design is sampled with probability proportional to a design being optimal \citep{thompson}---respectively.
ParEGO has been extended to the batch setting in various ways including: (i) MOEA/D-EGO~\citep{zhang_moead}, an algorithm that optimizes multiple scalarizations in parallel using MOEA/D~\citep{moead}, and (ii) $q$ParEGO~\citep{daulton2020ehvi}, which uses composite objectives with sequential greedy batch selection under different scalarization weights.
Information-theoretic methods, e.g., \citet{pesmo, pfes} have also garnered recent interest.

LaMOO \citep{zhao2021multiobjective} is a recent work that partitions the search space into ``good`` and ``bad`` regions and samples new designs from ``good`` regions using $q$EHVI or CMA-ES \citep{cmaes}. However, LaMOO-$q$EHVI relies on global GPs and is therefore prohibitively time-consuming with large evaluation budgets.
In addition, the authors propose to use rejection sampling to enforce that samples are from the, typically non-rectangular, "good" region, but rejection sampling is prohibitively time-consuming in high-dimensional search spaces (see Appendix~\ref{appdx:lamoo} for further discussion).
\subsubsection{High-dimensional Bayesian optimization}
Two popular approaches for high-dimensional BO are (1) mapping the high-dimensional inputs to a low-dimensional space via a random embedding~\citep{wang2016rembo,HeSBO19,letham2019re} and (2) exploiting additive structure~\citep{kandasamy15,gardner2017discovering}.
However, both families of methods require strong assumptions on the structure of the problem (low-dimensional linear or additive structure, respectively), and often perform poorly if the assumptions do not hold~\citep{eriksson2021high}.
This is especially problematic when optimizing multiple objectives since all objectives need to have the same assumed structure, which is unlikely in practice.
\citet{eriksson2021high} leverage a weaker assumption that the objective only depends on a small subset of the parameters and \citet{eriksson2021latencyaware} extended this approach to the multi-objective setting, but this approach
requires using computationally-demanding Markov Chain Monte Carlo methods for fitting the model, which is only feasible in the small data regime.
\subsubsection{Trust Region Bayesian Optimization}
Another popular method for high-dimensional BO is TuRBO \citep{eriksson2019turbo}, which performs BO in local trust regions (TRs) to avoid over-exploration. In contrast with \citep{zhao2021multiobjective} which uses non-rectangular "good" regions, TuRBO uses hyperrectangular TRs, where each TR $\mathcal{T}$ has a center point $\bm x_\text{center}$ and an edge-length $L \in [L_\text{min}, L_\text{max}]$.
Each TR maintains success and failure counters that record the number of consecutive samples generated from the TR that improved or failed to improve (respectively) the objective.
If the success counter exceeds a predetermined threshold~$\tau_{\text{succ}}$, the TR length is increased to $\min\{2L, L_\text{max}\}$ and the counter is reset to zero.
Similarly, after $\tau_{\text{fail}}$ consecutive failures the TR length is set to $L/2$ and the failure counter is set to zero.
Finally, if the length $L$ drops below a minimum edge length $L_\text{min}$, the TR is terminated and a new TR is initialized.

In contrast with aforementioned methods, TuRBO makes no strong assumptions about the objectives.
Although TuRBO has been extended to handle black-box constraints \citep{eriksson2021scalable}, to our knowledge, all existing TR-based BO methods target single-objective optimization.
In addition, TuRBO does not pass information between TRs, which results in an inefficient use of the evaluation budget; these methods have not observed significant improvement from using multiple TRs.
Lastly, even though optimization is restricted to a local TR, TuRBO fits GP models to the entire history of data collected by a single TR which can lead to poor scalability in settings where TRs restart infrequently.
\subsection{Issues with Scalarized TuRBO}
\begin{figure*}[!ht]
    \centering
    \includegraphics[width=0.96\textwidth]{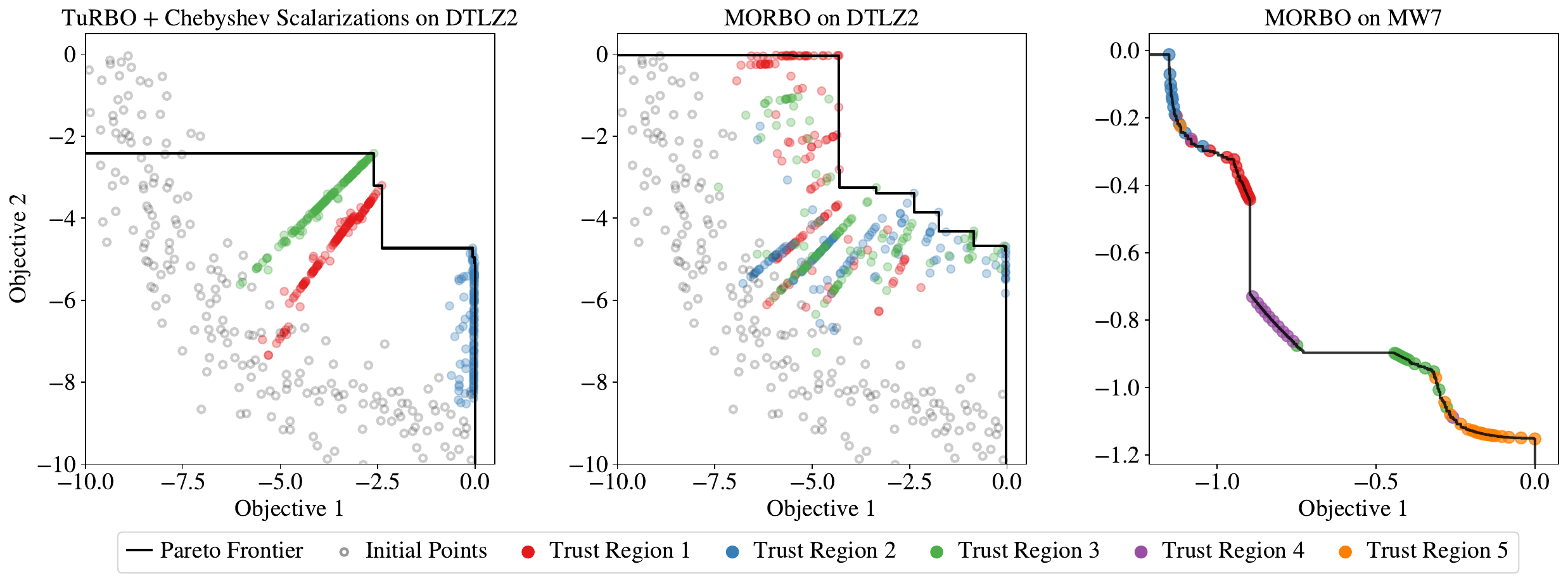}
    \caption{
        Objective values achieved on a $2$-objective DTLZ2 function with $d=100$ after $600$ evaluations, batch size $50$, and $3$ TRs.
        The scatter plot illustrates the search behavior.
        The grey circles indicate the initial space-filling design, which is the same for both methods. The other marker shapes and colors indicate which of the $3$ TRs obtained a given solution.
        The black line indicates the approximate Pareto frontier identified by each method.
        (Left) A straightforward extension of TuRBO where each TR optimizes a random Chebyshev scalarization of the objectives does not explore the trade-offs between the objectives because the TRs are rarely terminated under this approach, which leads only a few scalarizations being used.
        (Center) In contrast, \ALG employs a center selection strategy that actively targets under-explored regions of the Pareto frontier and uses a hypervolume-based acquisition function that is known to reward to high quality Pareto frontiers~\citep{zitzler03, Couckuyt14, yang2019} and explores the entirety of the PF.
        (Right) \ALG can discover disconnected regions of global PF on the MW7 function ($d=10$, with $2$ constraints) by using $5$ TRs to locally optimize disjoint regions of PF collaboratively, in parallel. This is stark contrast with TuRBO with Chebyshev scalarizations which the left plot shows yield approximate Pareto frontiers with poor coverage and diversity, even when the true PF is connected and simple.
    }
    \label{fig:intro_fig}
\end{figure*}

Since ParEGO is a well-established method (in low-dimensional settings) that optimizes random Chebyshev scalarizations, a reasonable approach would be to extend TuRBO to the MOO setting by using multiple TRs in parallel where each TR optimizes a different random Chebyshev scalarization of the objectives.
However, as we demonstrate in the left subplot of Figure~\ref{fig:intro_fig}, this approach results in a PF with very poor coverage. This is because a single scalarization is used for the lifetime of each TR in order to maintain a stable objective. Optimizing a single scalarization per trust region often leads to better solutions with respect to that scalarization than optimizing the entire PF using a hypervolume-based acquisition functions, which requires exploration of different objective trade-offs. However, if TRs are not restarted frequently (e.g. because TuRBO continues to find better solutions with respect to that scalarization), only a small number of scalarizations will be used, which can lead to poor coverage of the PF. As shown in Figure 2, we observe that MORBO yields PFs with better coverage (diversity of trade-offs). In addition, the TRs in TuRBO are independent; they do not pass information about evaluated designs and observations, and they do not collaboratively aim to optimize the global PF---rather, they act in isolation to optimize their own objectives. Together, this leads to an inefficient use of the sample budget.

\section{MORBO}
\label{sec:hdbpomo}
We now introduce \ALG, a \emph{collaborative} multi-TR approach for constrained high-dimensional multi-objective BO.
Rather than following TuRBO's approach of employing multiple independent TRs, \ALG shares observations across TRs to provide each TR with all available information about the objectives and constraints relevant for local optimization in the TR. Moreover, \ALG further departs from TuRBO by (1) selecting TR center points in a coordinated fashion to encourage identifying Pareto frontiers with good coverage, (2) choosing new candidate designs by collaboratively optimizing a shared global utility, and (3) employing local models to reduce computational complexity and improve scalability in large data regimes.
As shown in the center plot of Figure~\ref{fig:intro_fig}, \ALG identifies a high quality PF with much better coverage than the aforementioned simple TuRBO extension.
For the remainder of this section, we describe the core components of \ALG, which are also summarized in Algorithm~\ref{algo}.
%

\begin{algorithm*}[!ht]
    \DontPrintSemicolon
    \KwIn{Objective functions $f$, Number of trust region $n_\text{TR}$, Initial trust region length $L_\text{init}$, Maximum trust region length $L_{\max}$, Minimum trust region length $L_{\min}$.}
    \KwOut{Approximate Pareto frontier $\mathcal P_n$}
    Evaluate an initial set of points and initialize the trust regions $\mathcal T_1, ..., \mathcal T_{n_\text{TR}}$ using the center selection procedure described in Section~\ref{sec:center_selection} and mark center points as unavailable for other trust regions. \\
    $X_0 \leftarrow \emptyset{}, Y_0 = \emptyset{}, t \leftarrow 1$\\
    \While{budget not exhausted}{
        Fit a local model within each trust region. \\
        Select $q$ candidates using the sequential greedy \HVI{} procedure described in Section~\ref{sec:batch_selection}.\\
        Evaluate candidates on the true objective functions and obtain new observations.\\
        \For{$j=1,..., n_\text{TR}$}{
            Update trust regions with new observations as described in Section \ref{sec:hdbpomo}.\\
            Increment success/failure counters as described in Section \ref{sec:hdbpomo} for observations from $T_j$.\\
            Update edgelength $L_j$ for $\mathcal T_j$.\\
            \If{$L_j < L_\text{min}$}{
                Terminate $\mathcal T_j$.\\
                Fit GP to restart points $\mathcal D_{t-1} = (X_{t-1},Y_{t-1})$: $\bm f_{t-1} \sim P(\bm f | \mathcal D_{t-1})$.\\
                Sample $\bm\lambda\sim S_+^{M-1}$ and $\tilde{\bm f}_{t-1} \sim P(\bm f | \mathcal D_{t-1})$, where $S_+^{M-1} = \{\bm w\in \mathbb R_+^M : ||\bm w||_2 = 1\}$.\\
                $\bm x_t \leftarrow \argmax_{\bm x \in \mathcal X} s_{\bm\lambda}[\tilde{\bm f}_{t-1}(\bm x)]$, where $s_{\bm\lambda}[\bm y] = \min_m (\max(\frac{y_m}{\lambda_m}, 0))^M$ and $\cdot_i$ denotes the $i^\text{th}$ element.\\
                Evaluate $\bm x_t$ on the true objective functions and obtain new observation $\bm y_t$.\\
                Reinitialize $\mathcal T_j$ with edgelength $L_\text{init}$ centered at the $\bm x_t$.\\
                Set $X_t \leftarrow X_{t-1} \cup \{\bm x_t\}, Y_t \leftarrow Y_{t-1} \cup \{\bm y_t\}$, $t \leftarrow t+1$.

            }
            Update center to the available point with maximum \HVC{} (globally if $\mathcal T_j$ was terminated otherwise within $\mathcal T_j$).
        }
    }
    \Return{Approximate PF across observed function values}.
    \caption{Summary of \ALG}
\label{algo}
\end{algorithm*}
\subsection{Collaborative Batch Selection via Global Utility Maximization}\label{sec:batch_selection}
Maximizing hypervolume improvement (\HVI{}) has been shown to produce high-quality and diverse PFs~\citep{emmerich2006}.
Given a reference point, the hypervolume improvement from a set of points is the increase in \HV{} when adding these points to the previously selected points.
Expected HVI (EHVI) is a popular acquisition function that integrates \HVI{} over the GP posterior.
However, maximizing EHVI directly requires re-computing the GP posterior and sampling from it in each gradient step, which becomes prohibitively slow as the number of objectives (and constraints) and in-sample data points increases.

To allow scalability to large batch sizes $q$, we instead use Thompson sampling (TS) to draw $q$ posterior samples from the GP and optimize \HVI{} under each realization. This approach can be viewed as a single-sample approximation of EHVI \citep{daulton2021nehvi}.
We select $q$ points $\bm x_1, ..., \bm x_q$ for the next batch in a \emph{sequential greedy} fashion and condition upon the previously selected points in the batch by computing the HVI with respect to the current PF $\mathcal P$.
In particular, to select the $i^\text{th}$ point from a set of $r$ candidate points $\hat{\bm x}_1, \ldots, \hat{\bm x}_r$ we draw a sample from the joint posterior over $\bm f(\{\bm x_{1}, \ldots, \bm x_{i-1}\} \cup \{\hat{\bm x}_1, \ldots, \hat{\bm x}_r\})$, which yields the realization $\{\tilde{\bm f}(\bm x_{1}), \ldots, \tilde{\bm f}(\bm x_{i-1}), \tilde{\bm f}(\hat{\bm x}_1), \ldots, \tilde{\bm f}(\hat{\bm x}_r)\}$.
We select the $i^\text{th}$ point as the candidate point that maximizes the HVI jointly with the realizations $\tilde{\bm f}(\bm x_{1}), \ldots, \tilde{\bm f}(\bm x_{i-1})$ of the previously selected points as shown in Figure~\ref{fig:hvi}.
Conditioning on the previously selected points and computing the HVI under a sample from the joint posterior over the previously selected points and the discrete set of candidates leads to more diverse batch selection compared to selecting each point independently. Moreover, this approach effectively lets TRs collaboratively maximize the global HVI utility function.
Using this global utility, an individual TR considers the iteration a success if at least one proposed candidate improves the global HV and a failure otherwise.

Another benefit of HV-based acquisition functions is that they naturally provide utility values for set of points, which enables the TRs to target different parts of the PF. 
This is particularly appealing in settings where the PF may be disjoint or may require exploring different parts of the search space. As shown in the right plot of Figure~\ref{fig:intro_fig}, \ALG recovers diverse regions of a disconnected PF.
Lastly, we note that this batch selection strategy also allows to straightforwardly implement \emph{fully asynchronous} optimization, where evaluations are dispatched to different ``workers'' and new candidates are generated whenever there is capacity in the worker pool.
In the asynchronous setting, success/failure counters and TRs can be updated after every $q$ observations are received, and intermediate observations can immediately be used to update the local models.
\subsection{Coordinated Trust Region Center Selection}\label{sec:center_selection}
In (constrained) single-objective optimization, previous work centers the local TR at the best (feasible) observed point.
However, in the multi-objective setting, there is typically no single best solution.
Assuming noise-free observations, \ALG selects the center to be the feasible point on the PF with maximum hypervolume contribution (\HVC{})~\citep{Beume2007,loshchilov2011}.
If there is no feasible point, \ALG chooses the point with the smallest total constraint violation (see Appendix~\ref{appdx:constraint_handling} for details on center selection with constraints).
Given a reference point, the \HVC{} of a point on the PF is the reduction in HV if that point were to be removed; that is, the HVC of a point is its exclusive contribution to the PF.
Centering a TR at the point with maximal HVC collected by that TR promotes coverage across the PF, as points in crowded regions will have lower contribution.
\ALG selects TR centers based on their HVCs in a sequential greedy fashion, excluding points that have already been selected as the center for another TR.

\subsection{Local Modeling}\label{sec:local_modeling}
Most BO methods use a single global GP model, often with a stationary kernel (e.g. Mat\'ern-$5/2$) using automatic relevance determination (ARD) fitted to all observations collected so far.
While a global model is necessary for most BO methods, \ALG only requires each model to be accurate within the corresponding TR.
To increase scalability, we employ local modeling where we only include the observations contained within a local modeling hypercube with edge length~$2L$. The motivation for using the observations from a slightly larger hypercube is to improve the model close to the TR boundary.

In previous trust region BO works \citep{eriksson2019turbo, eriksson2021scalable, wan2021think}, each TR uses a GP that is fitted to the all observations collected by that TR (rather than only a set of local observations in or near the TR), which leads to scalability issues due to the cubic time complexity of GP inference if the TR collects many observations.
In addition, fitting a GP solely to data collected by a single TR ignores observations collected by other TRs and makes inefficient use of the sampling budget.
In contrast, \ALG shares observations across TRs and employs local models, where models are fit to all observations within a hypercube with edge length~$2L$. This significantly reduces the computational cost since exact GP fitting scales cubically with the number of data points. Under limited assumptions on the distribution of data across TRs, using local models results in speedups of $O(n_\text{TR}^2 / \eta^3)$, where $\eta$ is the average number of TR modeling spaces a data point resides in. Empirically, we demonstrate (see Figure~\ref{fig:tr_traces} in Appendix~\ref{appdx:additional_results}) that $\eta < 1$ as the optimization progresses and the TRs shrink, and we find that this translates into speedups of two orders of magnitude relative to global modeling as shown in Appendix~\ref{appdx:wall_time_comparisons}. See Appendix~\ref{appdx:complexity_local} for more details on the complexity.
\subsection{Re-initialization Strategy}
Although \ALG performs local optimization within a TR, we ensure global optimization by re-initializing TRs using a
principled  
technique based on hypervolume scalarizations~\citep{zhang2020random}. A HV scalarization is defined as $s_{\bm\lambda}[\bm y] = \min_m (\max(\frac{y_m}{\lambda_m}, 0))^M$, where $\cdot_m$ denotes the $m^\text{th}$ component~\citep{zhang2020random}. Let $\mathcal D_{t-1} = (X_{t-1}, Y_{t-1})$ be the set of previous re-initialization (restart) points $X_{t-1} = \{\bm x_i\}_{i=1}^{t-1}$ and corresponding observations $Y_{t-1} = \bm f(X_{t-1})$, where $X_0 = \emptyset{}$ and $Y_0 = \emptyset{}$. Given $\mathcal D_{t-1}$, we determine the center point $x_t$ of the new TR by maximizing a random HV scalarization of the objectives under a posterior sample from a global GP posterior conditioned on $D_{t-1}$: $\tilde{\bm f} \sim P(\bm f|\mathcal D_{t-1})$. This ensures that TRs are initialized in diverse parts of the objective space and yields a global optimization performance guarantee (Section~\ref{sec:theory}).


\section{Theoretical Analysis}
\label{sec:theory}
We analyze the performance of \ALG{} in terms of its cumulative HV regret. The instantaneous HV regret $R(\mathcal P_t)$ after $t$ TR restarts is defined as the difference in HV dominated by the true Pareto frontier $\mathcal P^*$ and the approximate Pareto frontier $\mathcal P_t$: $R(\mathcal P_t) = \HV{}(\mathcal P^*) -  \HV{}(\mathcal P_t).$ The (cumulative) HV regret after $T$ restarts is the sum of the instantaneous regret over all restarts: $R_T = \sum_{t=1}^T R(\mathcal P_t)$. First, we show that a TR will only evaluate a finite number of samples before restarting.
\begin{lemma}
\label{lemma:restarts}
Let $\bm f \in [0, B]^M$, and assume that \ALG{} only considers a newly evaluated sample to be an improvement (for updating the corresponding TR's success and failure counters) if it increases the HV by at least $\delta \in \mathbb R^+$ and assume that success counter threshold $\tau_\text{succ} = \infty$.\footnote{As stated in Appendix \ref{appdx:experimentdetails}, we use $\tau_\text{succ} = \infty$ in all of our experiments.} Then each TR will only evaluate a finite number of samples.
\end{lemma}
The proof is given in Appendix~\ref{appdx:proofs}. Having established that TRs only evaluate a finite number of designs, we now bound the hypervolume regret with respect to the number of restarted TRs.
The bound leverages the kernel-dependent maximum information gain $\gamma_T$---which measures the decrease in uncertainty after $T$ observations ---and is commonly used to analyze regret in BO \citep{ucb}.
\begin{theorem}
\label{thm:hv_regret}
Let $\bm f \in [0,B]^M$ for $B>0$ and let each component $f^{(m)}$ for $m=1, ..., M$ follow a Gaussian distribution with marginal variances $\sigma \leq 1$ and independent observation noise $\epsilon_m \sim \mathcal N(0, \sigma_m^2)$ such that $\sigma_m^2 \leq \sigma^2 \leq 1$. Let $\mathcal P_t$ denote the Pareto frontier over $\bm f(X_t)$, where $X_t$ is the set of TR re-initialization points after $t$ TRs have been restarted.
Suppose further that the conditions of Lemma~\ref{lemma:restarts} hold.
Then, the cumulative hypervolume regret $R_T$ of \ALG after $T$ restarts is bounded by:
$$R_T \leq M^2(\sqrt{2e\pi}B/2)^M \sqrt{d\gamma_T T\ln(T)}.$$
\end{theorem}
Up to logarithmic terms, this regret bound is on the order of $\tilde{\mathcal O}(\sqrt{T})$.
This bound is significant because, to our knowledge, \citet{zhang2020random} is the only other work to bound the HV regret of multi-objective BO algorithms.
This makes \ALG{} the first sample-efficient large-scale, MOO algorithm with bounded regret.
The proof, given in Appendix~\ref{appdx:proofs}, leverages the hypervolume regret bound from \citet{zhang2020random}.
However, our regret bound is with respect to the number of restart points (rather than evaluations)---a difference that can be viewed as a cost of focusing on large-scale problems which BO with global GPs cannot address.
Moreover, our regret analysis in terms of the number of restarts is similar to the convergence guarantees of gradient-based TR optimization methods~\citep{Yuan_areview} and can be viewed as a multi-objective analogue of the performance guarantees of recent single-objective BO-based TR methods \citep{wan2021think}.

\section{Experiments}
\label{sec:Experiments}
We evaluate \ALG on an extensive suite of benchmarks with various numbers of input parameters ($d$), objectives ($M$), and constraints ($V$).
In Appendix~\ref{appdx:baby_problems}, we consider a vehicle ($d=5$) and a welded beam ($d=4$, $V=4$) design problem to show that \ALG{} is competitive with other algorithms on problems it was not designed for.
We consider three challenging real-world problems: a trajectory planning problem ($d=60$), a problem of designing optical systems for AR/VR applications ($d=146$), and an automotive design problem ($d=222, V=54$) .
In addition, we evaluate \ALG{} on DTLZ3, DTLZ5, and DTLZ7 problems with $2$/$4$ objectives ($6$ problems in total) in Appendix \ref{appdx:additional_results}.

We compare \ALG to multi-objective BO methods ($q$NEHVI, $q$ParEGO, TS-TCH, TSEMO, DGEMO, MOEA/D-EGO), recent work leveraging search space partitioning (LaMOO-CMAES, LaMOO-$q$NEHVI), a widely used evolutionary algorithm (NSGA-II), and Sobol---a quasi-random baseline where designs are sampled from a scrambled Sobol sequence \citep{owen2003quasi} (see Appendix~\ref{appdx:experimentdetails} for more details on the methods).
\ALG is implemented using BoTorch~\citep{balandat2020botorch} and the code will be made publicly available soon.
We run all methods for $20$ replications and initialize them using the same quasi-random initial points for each replication.
We use the same hyperparameters for \ALG on all problems and conduct analyze the sensitivity of \ALG to its hyperparameters in Figure~\ref{fig:ablation_study}.
See Appendix~\ref{appdx:experimentdetails} for details on the experiment setup.
All experiments used a Tesla V100 SXM2 GPU (16GB RAM).
\begin{figure*}[!ht]
    \centering
    \includegraphics[width=0.96\textwidth]{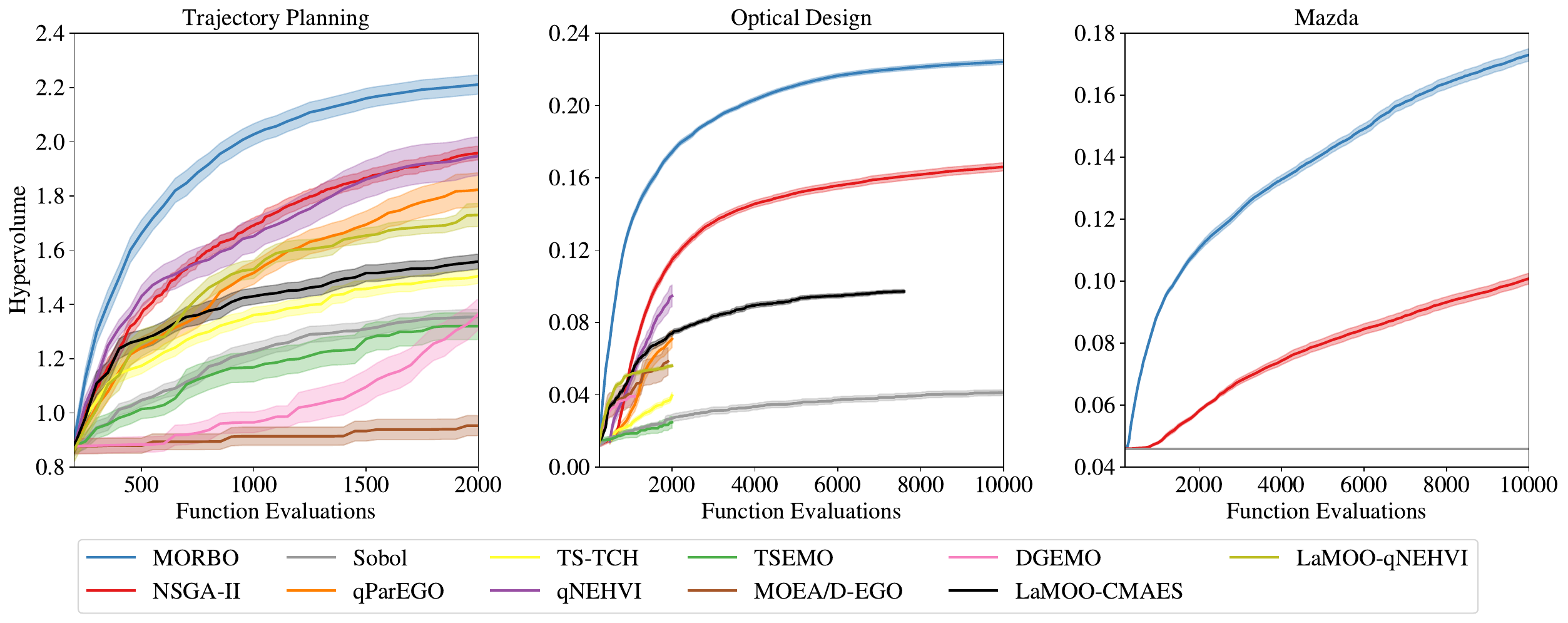}
    \caption{
        (Left) \ALG outperforms other methods on the trajectory planning problem ($d=60$).
        (Middle) Illustration of the results on the Optical design problem ($d=146$). NSGA-II performs better than the BO baselines but is not competitive with \ALG.
        (Right) \ALG shows compelling performance on the Mazda vehicle design problem ($d=222$) with $54$ black-box constraints. For all plots, we show the mean and one standard error of the mean over 20 replications.
    }
    \label{fig:real_world_experiments}
\end{figure*}
\subsection{Large-Scale Real-World Problems}
\paragraph{Trajectory Planning}
\label{sec:real_world_problems}
We consider a trajectory planning problem similar to the rover trajectory planning problem considered in~\citep{wang2018batched}.
As in the original problem, the goal is to find a trajectory that maximizes the reward when integrated over the domain.
The trajectory is determined by fitting a B-spline to $30$ design points in the 2-objective plane, which yields a $60$-dimensional optimization problem.
In this experiment, we constrain the trajectory to begin at the pre-specified starting location, but we do not require it to end at the desired target location.
In addition to maximize the reward of the trajectory, we also minimize the distance from the end of the trajectory to the intended target location.
Intuitively, these two objectives are expected to be competing because reaching the exact end location may require passing through areas with lower associated reward.
The results from {$2$,$000$} evaluations using batch size $q=50$ and $200$ initial points are presented in Figure~\ref{fig:real_world_experiments}, which shows that \ALG performs the best and even state-of-the-art methods such as $q$NEHVI do not out perform NSGA-II.

\paragraph{Optical design problem}
We consider the problem of designing an optical system for an augmented reality (AR) see-through display.
This optimization task has $146$ parameters describing the geometry and surface morphology of multiple optical elements in the display stack.
Several objectives are of interest in this problem, including display efficiency and display quality.
Each evaluation of these metrics requires a computationally intensive physics simulation 
that takes several hours to run.
In this benchmark, the task is to explore the Pareto frontier between display efficiency and display quality (both objectives are normalized w.r.t. the reference point).
We consider $250$ initial points, batch size $q=50$, and a total of {$10$,$000$} evaluations.
This is out of reach for the other BO baselines due to runtime considerations, and so we run $q$NEHVI, $q$ParEGO, TS-TCH, TSEMO, MOEA/D-EGO, for {$2$,$000$} evaluations and DGEMO for {$1$,$000$} evaluations.
We were only able to run LaMOO-CMAES for $7,600$ evaluations before it overflowed GPU memory.
Figure~\ref{fig:real_world_experiments} shows that \ALG achieves substantial improvements in sample efficiency compared to NSGA-II.
Furthermore, observe that no other baselines are competitive with NSGA-II except in the very small sample regime (less than $500$ evaluations).

\begin{figure*}[!ht]
    \centering
    \includegraphics[width=0.96\textwidth]{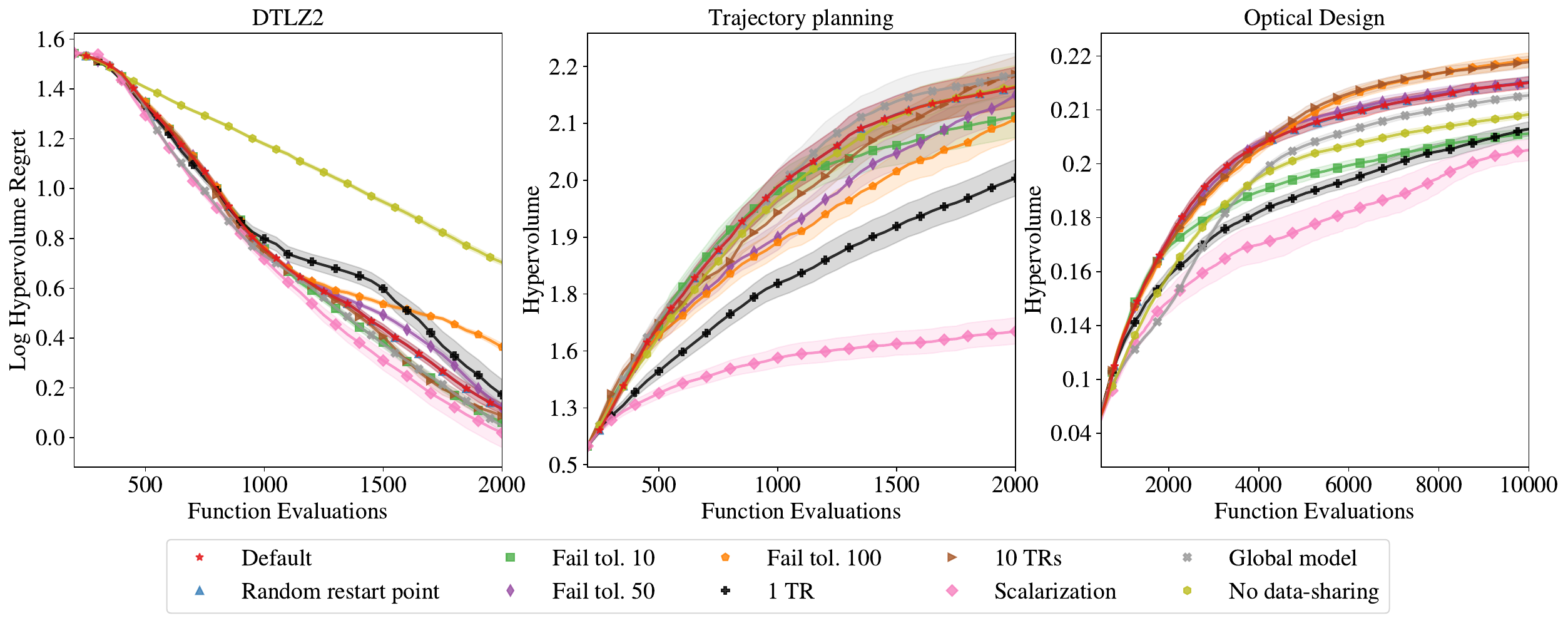}
    \caption{
        We investigate the sensitivity of \ALG with respect to its hyperparameters.
        We observe that using multiple TRs performs significantly better than using a single TR and that data-sharing and the use of a hypervolume based acquisition function are important components of \ALG.
    }
    \label{fig:ablation_study}
\end{figure*}

\paragraph{Mazda vehicle design problem}
We consider the $3$-car Mazda benchmark problem~\citep{kohira2018proposal}.
This challenging MOO problem involves tuning $222$ decision variables that represent the thickness of different structural parts.
The goal is to minimize the total vehicle mass of the three vehicles (Mazda CX-$5$, Mazda $6$, and Mazda $3$) as well as maximizing the number of parts shared across vehicles.
Additionally, there are $54$ black-box output constraints (evaluated jointly with the two objectives) that enforce that designs meet performance requirements such as collision safety standards.
This problem is, to the best our knowledge, the largest MOO problem considered by any BO method and requires fitting $56$ GP models to the objectives and constraints.
The original problem underlying the Mazda benchmark was solved on what at the time was the world's fastest supercomputer and took around $3$,$000$ CPU years to compute~\citep{oyama2017mazda}.
We consider a budget of {$10$,$000$} evaluations using batches of size $q=50$ and $300$ initial points.

Figure~\ref{fig:real_world_experiments} demonstrates that \ALG clearly outperforms the other methods.
A feasible design satisfying the black-box constraints was provided to all methods for all replications as part of the initial $300$ design points.
However, in subsequent evaluations Sobol did not find another feasible design, illustrating the challenge of satisfying the $54$ constraints.
While NSGA-II made progress from the initial feasible solution, it is not competitive with \ALG.
NSGA-II and Sobol are the only applicable baselines because standard multi-objective BO methods are impractically slow with $56$ \emph{global} GPs and LaMOO does not support black-box constraints.
\subsection{Ablation study}\label{sec:ablation}
Finally, we study the sensitivity of \ALG with respect to the number of TRs ($n_\text{TR}$), the failure tolerance ($\tau_\text{fail}$), and sharing observations across TRs, local modeling, HVI acquisition function, and the re-initialization strategy.
Using several TRs allows \ALG to explore different parts of the search space that potentially contribute to different parts of the Pareto frontier.
The failure tolerance controls how quickly each TR shrinks:
A large $\tau_\text{fail}$ leads to slow shrinkage and potentially too much exploration, while a small $\tau_\text{fail}$ may cause each TR to shrink too quickly and not collect enough data.
\ALG uses $5$ TRs and $\tau_\text{fail} = \max(10, \frac{d}{3})$ by default, similar to what is used by~\citet{eriksson2019turbo}.

We consider the DTLZ2 problem ($d=100$, $M=2$), the trajectory planning problem ($d=60$, $M=2$), and the optical design problem ($d=146$, $M=2$).
Figure~\ref{fig:ablation_study} shows that \ALG with the default settings performs well on all three problems.
We observe that multiple TRs and the HVI acquisition function are important as neither a single TR nor a Chebyshev scalarization performs well.
The performance of \ALG is robust to the choice of failure tolerance except for on the optical design problem where using a value of $10$ is clearly worse than the default and causes the TRs to shrink too quickly.
Not sharing data between TRs results in inferior results on the DTLZ2 and optical design problems.
While using a global GP model achieves good results on the DTLZ2 and trajectory planning problems, it does not perform as well on the optical design problem.
A global GP also comes at a high computational cost.
Using a global GP, running \ALG with a budget of {$10$,$000$} evaluations on the optical design problem required $30$ hours of computational overhead, whereas \ALG did {$10$,$000$} evaluations in less than an hour using local models.
Lastly, we find consistently strong performance for both our default HV scalarization-based re-initialization strategy and a strategy that selects a new design at random (denoted as "Random restart points").
The former allows us to bound \ALG's regret.

\section{Discussion}
\label{sec:Discussion}
We proposed \ALG, an algorithm for multi-objective BO over high-dimensional search spaces.
By using a coordinated, collaborative multi-trust-region approach with scalable local modeling, \ALG scales gracefully to high-dimensional problems and high-throughput settings.
In a comprehensive experimental evaluation, we showed that \ALG allows us to \emph{effectively tackle important real-world problems that were previously out of reach for existing BO methods}.
We showed that \ALG achieves substantial improvements in sample efficiency compared to existing state-of-the-art methods such as evolutionary algorithms.
Due to the lack of alternatives, NSGA-II has been the method of choice for many practitioners, and we expect \ALG to provide practitioners with significant savings in terms of time and resources across the many disciplines that require solving challenging optimization problems.

However, there are some limitations to our method.
Although \ALG can handle a large number of black-box constraints, using hypervolume-based acquisition means the computational complexity scales poorly with the number of objectives.
Furthermore, \ALG is optimized for the large-batch high-throughput setting and other methods may be more suitable for and achieve better performance on low-dimensional problems with small evaluation budgets.


\newpage
\bibliography{daulton_446}

\appendix
\onecolumn

\section{Details on Batch Selection}
\FloatBarrier
\label{appdx:batch_selection}
\begin{figure}[ht]
    \centering
    \includegraphics[width=0.33\textwidth]{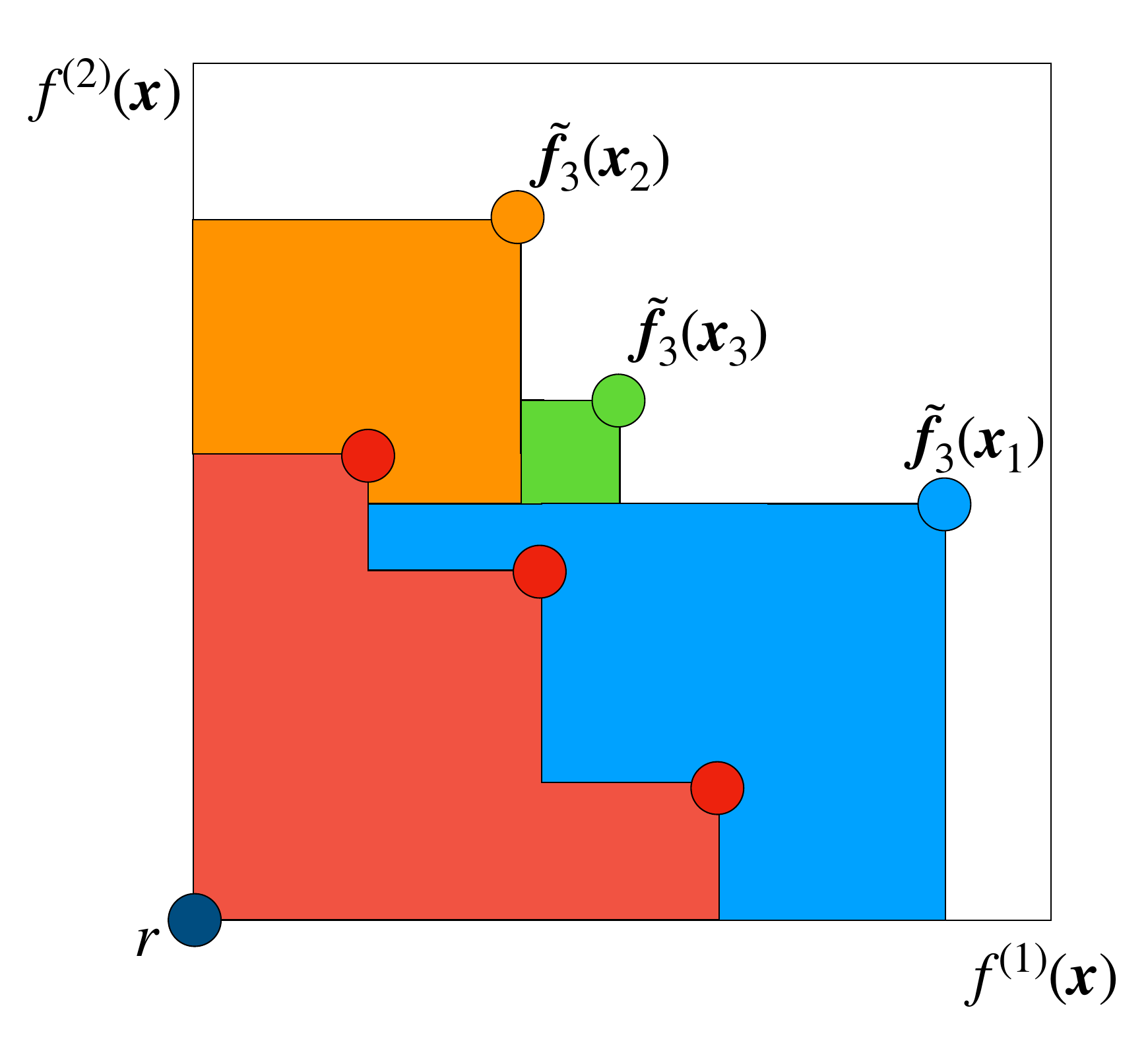}
    \caption{
        A visualization of our batch selection using \HVI{} with $q=4$.
        The red points represent the current PF.
        Blue, orange, and green points show the function values for the $3$ selected points under the next posterior sample. 
        To select the $4$th point, the \HVI{} of each candidate is evaluated jointly with the red, blue, orange, and green points.
    }
    \label{fig:hvi}
\end{figure}
As discussed in Section~\ref{sec:background}, over-exploration can be an issue in high-dimensional BO because there is typically high uncertainty on the boundary of the search space, which often results in over-exploration. 
This is particularly problematic when using continuous optimization routines to find the maximizer of the acquisition function since the global optimum of the acquisition function will often be on the boundary, see~\cite{oh2018bock} for a discussion on the ``boundary issue'' in BO.
While the use of trust regions alleviates this issue, this boundary issue can still be problematic, especially when the trust regions are large.

To mitigate this issue of over-exploration, we use a discrete set of candidates by perturbing randomly sampled Pareto optimal points within a trust region by replacing only a small subset of the dimensions with quasi-random values from a scrambled Sobol sequence.
This is similar to the approach used by~\citet{eriksson2021scalable} which proved crucial for good performance on high-dimensional problems.
In addition, we also decrease the perturbation probability~$p_n$ as the optimization progresses, which \citet{Regis2013} found to improve optimization performance. 
The perturbation probability~$p_n$ is set according to the following schedule:
$$ p_n = p_0 \bigg[1 - 0.5\frac{\log n'}{\log b}\bigg],$$
where $n_0$ is the number of initial points, $n_f$ is the total evaluation budget, $p_0 = \min\{\frac{20}{d}, 1\}$, $b = n_f - n_0$, and $n' = \min\{\max\{n - n_0, 1\}, b\}$.

Given a discrete set of candidates, \ALG{} draws samples from the joint posterior over the function values for the candidates in this set and the previously selected candidates in the current batch, and selects the candidate with maximum \HVI{} across the joint samples. 
This procedure is repeated to build the entire batch.\footnote{In the case that the candidate point does not satisfy that satisfy all outcome constraints under the sampled GP function, the acquisition value is set to be the negative constraint violation.} 
Using standard Cholesky-based approaches, exact posterior sampling has complexity that is cubic with respect to the number of test points and therefore is only feasible for relatively small discrete sets. 
\FloatBarrier
\subsection{RFFs for fast posterior sampling}
\label{appdx:rffs}
While asymptotically faster approximations than exact sampling exist; see \citet{pleiss2020fast} for a comprehensive review, these methods still limit the candidate set to be of modest size (albeit larger), which may not do an adequate job of covering a the entire input space.
Among the alternatives to exact posterior sampling, we consider using Random Fourier Features (RFFs) \citep{rahimi_rff}, which provide a deterministic approximation of a GP function sample as a linear combination of Fourier basis functions. 
This approach has empirically been shown to perform well with Thompson sampling for multi-objective optimization~\citep{tsemo}. 
The RFF samples are cheap to evaluate and which enables using much larger discrete sets of candidates since the joint posterior over the discrete set does not need to be computed. 
Furthermore, the RFF samples are differentiable with respect to the new candidate $\bm x$, and \HVI{} is differentiable with respect to $\bm x$ using cached box decompositions~\citep{daulton2021nehvi}, so we can use second-order gradient optimization methods to maximize \HVI{} under the RFF samples.

We tried to optimize these RFF samples using a gradient based optimizer, but found that many parameters ended up on the boundary, which led to over-exploration and poor BO performance.
In an attempt to address this over-exploration issue, we instead consider continuous optimization over axis-aligned subspaces which is a continuous analogue of the discrete perturbation procedure described in the previous section. 
Specifically, we generate a discrete set of candidates points by perturbing random subsets of dimensions according to $p_n$, as in the exact sampling case. 
Then, we take the top $5$ initial points with the maximum \HVI{} under the RFF sample. 
For each of these best initial points we optimize only over the perturbed dimensions using a gradient based optimizer. 

\begin{figure}[h]
    \includegraphics[width=\textwidth]{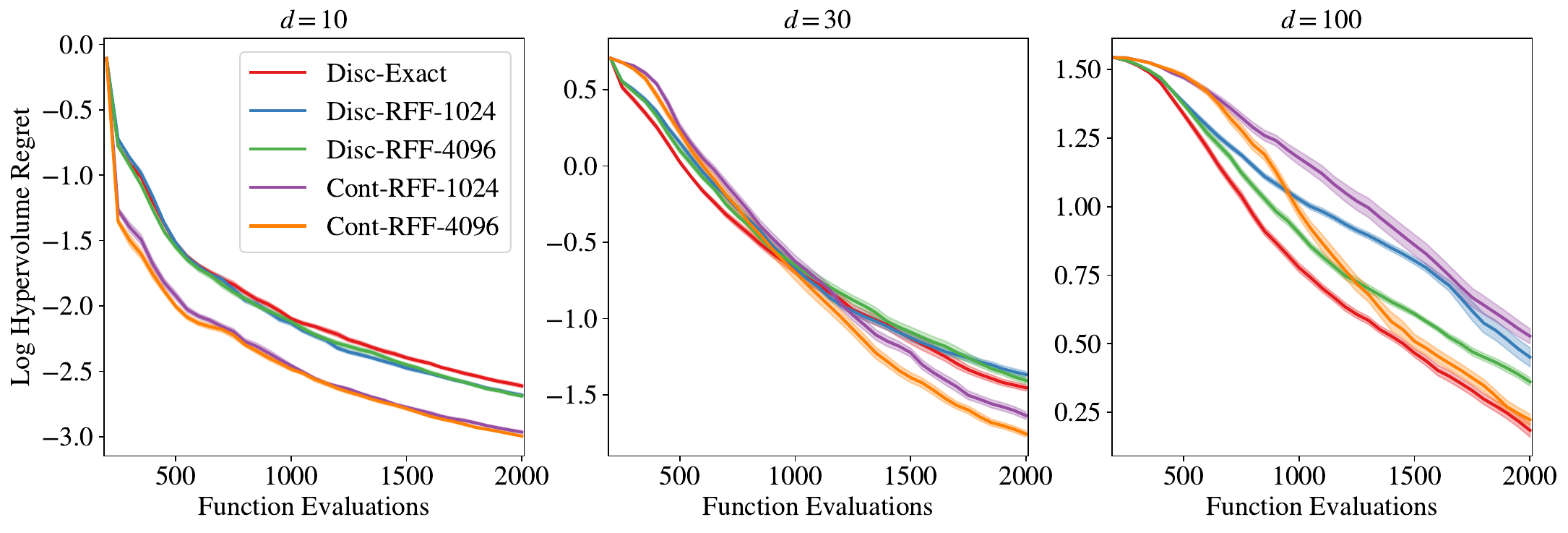}
    \caption{
        \label{fig:rff_vs_exact_ts} Optimization performance under various Thompson sampling approaches on DTLZ2 test problems with $2$ objectives and various input dimensions $d \in \{10,30,100\}$. 
        Disc-Exact uses exact samples from the joint posterior over a discrete set of $4096$ points. Disc-RFF-$1024$ and Disc-RFF-$4096$ evaluate approximate sample paths (RFFs) over a discrete set of $4096$ points with $1024$ and $4096$ basis functions, respectively. 
        Cont-RFF-$1024$ and Cont-RFF-$4096$ use L-BFGS-B with exact gradients to optimize RFF draws along a subset of the dimensions (see in Appendix \ref{appdx:rffs} for details) using $1024$ and $4096$ basis functions, respectively.
    }
\end{figure}

Figure \ref{fig:rff_vs_exact_ts} shows that the RFF approximation with continuous optimization over axis-aligned subspaces works well on for $D=10$ on the DTLZ2 function, but the performance degrades as the dimensionality increases.
Thus, the performance of \ALG can likely be improved on low-dimensional problems by using continuous optimization; we used exact sampling on a discrete set for all experiments in the paper for consistency.
We also see that as the dimensionality increases, using RFFs over a discrete set achieves better performance than using continuous optimization.
In high-dimensional search spaces, we find that exact posterior sampling over a discrete set achieves better performance than using RFFs, which we hypothesize is due to the quality of the RFF approximations degrading in higher dimensions. 
Indeed, as shown in Figure~\ref{fig:rff_vs_exact_ts}, optimization performance using RFFs improves if we use more basis functions on higher dimensional problems ($4096$ works better than $1024$).

\section{Additional details of constraint handling in \ALG}
\label{appdx:constraint_handling}
If there are feasible points, the center is selected as the point with maximum HVC across the feasible Pareto frontier. If there are no feasible points, the center is selected to be the point with minimum total constraint violation (the sum of the constraint violations). A TR’s success counter is incremented if the TR center was feasible and the candidates generated from this TR improved the feasible hypervolume or if the TR center was infeasible and a candidate generated from this TR has lower total constraint violation than the TR center. 
\section{Proofs}
\label{appdx:proofs}
\begin{manuallemma}{4.1}
Let $\bm f \in [0, B]^M$, and assume that \ALG{} only considers a newly evaluated sample to be an improvement (for updating the corresponding TR's success and failure counters) if it increases the HV by at least $\delta \in \mathbb R^+$ and assume that success counter threshold $\tau_\text{succ} = \infty$.\footnote{As stated in Appendix \ref{appdx:experimentdetails}, we use $\tau_\text{succ} = \infty$ in all of our experiments.} Then each TR will only evaluate a finite number of samples.
\end{manuallemma}
\begin{proof}
First, note that The hypervolume of the true Pareto frontier $\mathcal P^*$ is bounded. Without loss of generality, if the reference point $\bm r = \bm 0$, then the $\HV(\mathcal P^*) \leq B^M.$ Suppose that a trust region evaluates an infinite number of samples. Then, the trust region has not had $1+\log_2 L_\text{init}-\log_2L_\text{min}$ streaks of $\tau_\text{fail}$ consecutive failures. Hence, the trust region has increased the hypervolume of the Pareto frontier over the previously evaluated designs by at least $\delta$ infinitely many times. Hence, the hypervolume over the previously evaluated designs is infinite. This is a contradiction.
\end{proof}

\begin{manualtheorem}{4.1}
Let $\bm f \in [0,B]^M$ for $B>0$ and let each component $f^{(m)}$ for $m=1, ..., M$ follow a Gaussian distribution with marginal variances $\sigma \leq 1$ and independent observation noise $\epsilon_m \sim \mathcal N(0, \sigma_m^2)$ such that $\sigma_m^2 \leq \sigma^2 \leq 1$. Let $\mathcal P_t$ denote the Pareto frontier over $\bm f(X_t)$, where $X_t$ is the set of TR re-initialization points after $t$ TRs have been restarted. 
Suppose further that the conditions of Lemma~\ref{lemma:restarts} hold. 
Then, the cumulative hypervolume regret $R_T$ of \ALG after $T$ restarts is bounded by:
$$R_T \leq M^2(\sqrt{2e\pi}B/2)^M \sqrt{d\gamma_T T\ln(T)}.$$
\end{manualtheorem}
\begin{proof}
From Lemma~\ref{lemma:restarts}, we have that each trust region will only evaluate a finite number of samples. Hence, as the number of evaluations goes to infinity, \ALG{} will terminate and select new initial center points for trust regions an infinite number of times. Our regret bound is in terms of the number of restart points.

Our proof follows that of \citet[Theorem 8]{zhang2020random}, but the final form of our bound holds for arbitrary $B$. Note that lines 13-19 in Algorithm~\ref{algo} correspond to \citet[Algorithm 1]{paria2020flexible} using Thompson sampling, where the only evaluations are the $t-1$ restart points. From \citet[Theorem 1]{paria2020flexible}, the scalarized Bayes regret of \citet[Algorithm 1]{paria2020flexible} using $L$--Lipschitz scalarizations is $O\big(LMd^{\frac{1}{2}}[\gamma_T T\ln(T)]^{\frac{1}{2}}\big)$. Since a hypervolume scalarization $s_{\bm\lambda}[\bm y]$ is $\mathcal O(B^MM^{1+M/2})$--Lipschitz \citep[Lemma 6]{zhang2020random}, we have that $L\leq B^MM^{1+M/2}$. From \citet[Proof of Theorem 8]{zhang2020random}, the hypervolume regret can be expressed by scaling the scalarized Bayes regret by a constant $c_M=\frac{\pi^\frac{M}{2}}{2^M \Gamma(\frac{M}{2}+1)}$ that depends on the number of objectives. Hence, we can bound the hypervolume regret as:
\begin{align*}
    R_T = \sum_{t=1}^T\HV(\mathcal P^*) - \HV(\mathcal P_t) &\leq c_M LMd^{\frac{1}{2}}[\gamma_T T\ln(T)]^{\frac{1}{2}}.
\end{align*}
Note that
\begin{align*}
    c_ML &\leq B^M M^{1+M/2}\frac{\pi^\frac{M}{2}}{2^M \Gamma(\frac{M}{2}+1)}
\end{align*}
From \citet[Theorem 1]{Li2007}, $\Gamma(x) > \frac{x^{x-\gamma}}{e^{x-1}}$, where $\gamma \approx 0.577$ is the Euler-Mascheroni constant. So,
$$\Gamma(M/2+1)
> \frac{(M/2+1)^{(M/2+1-\gamma)}}{e^{(M/2)}}
>\frac{M^{(M/2)}}{2e^{(M/2)}}.
$$
Hence,
$$\frac{1}{\Gamma(\frac{M}{2}+1)} < \frac{(2e)^{(M/2)}}{M^{(M/2)}}.$$
So,
\begin{align*}
    c_ML &\leq B^M M^{1+M/2}\frac{\pi^\frac{M}{2}}{2^M \Gamma(\frac{M}{2}+1)}\\
    &\leq B^M M\frac{(2e\pi)^\frac{M}{2}}{2^M }\\
    &\leq M\big(\sqrt{2e\pi}B/2\big)^M.
\end{align*}
So the cumulative regret bound is
\begin{align*}
    R_T &\leq c_M LMd^{\frac{1}{2}}[\gamma_T T\ln(T)]^{\frac{1}{2}}\\
    &\leq  M^2(\sqrt{2e\pi}B/2)^Md^{\frac{1}{2}}[\gamma_T T\ln(T)]^{\frac{1}{2}}.
\end{align*}
\end{proof}

\section{Details on Experiments}
\label{appdx:experimentdetails}

\subsection{Algorithmic details}
For \ALG{}, we use $5$ trust regions, which we observed was a robust choice in Figure~\ref{fig:ablation_study}.
Following~\citep{eriksson2019turbo}, we set $L_{\text{init}}=0.8$, $L_{\max} = 1.6$, and use a minimum length of $L_{\min} = 0.01$.
We use $4096$ discrete points for optimizing \HVI{} for the vehicle safety and welded beam problems, $2048$ discrete points on the trajectory planning and optical design problems, and $512$ discrete points on the Mazda problem.
Note that while the number of discrete points should ideally be chosen as large as possible, it offers a way to control the computational overhead of \ALG; we used a smaller value for the Mazda problem due to the fact that we need to sample from a total of $56$ GP models in each trust region as there are $54$ black-box constraints.
We use an independent GP with a a constant mean function and a Mat\'ern-$5/2$ kernel with automatic relevance detection (ARD) and fit the GP hyperparameters by maximizing the marginal log-likelihood (the same model is used for all BO baselines).

When fitting a model for \ALG, we include the data within a hypercube around the trust region center with edgelength $2L$. 
In the case that there are less than $N_m := \min\{250, 2d\}$ points within that region, we include the $N_m$ closest points to the trust region center for model fitting. 
The success streak tolerance is set to be infinity, which prevents the trust region from expanding; we find this leads to good optimization performance when data is shared across trust regions. 
For $q$NEHVI and $q$ParEGO, we use $128$ quasi-MC samples and for TS-TCH, we optimize RFFs with $500$ Fourier basis functions. 
All three methods are optimized using L-BFGS-B with $20$ random restarts.
For DGEMO, TSEMO, and MOEA/D-EGO, we use the default settings in the open-source implementation at \url{https://github.com/yunshengtian/DGEMO/tree/master}. 
Similarly, we use the default settings for NSGA-II the Platypus package (\url{https://github.com/Project-Platypus/Platypus}). 
We encode the reference point as a black-box constraint to provide this information to NSGA-II.

\subsubsection{LaMOO in High-Dimensional Search Spaces}
\label{appdx:lamoo}
For LaMOO methods, leverage the implementation of LaMOO available at \url{https://drive.google.com/drive/folders/1CMdg5iBdbKe3nkboIjiS998rnBEV09EB?usp=sharing}. We set the exploration parameter $C_p$ dynamically using the heuristic proposed by \citet{zhao2021multiobjective} to be 10\% of the hypervolume of the current Pareto frontier over the previously evaluated designs. We follow \citet{zhao2021multiobjective} and set the minimum leaf sample size to be $10$.

\citet{zhao2021multiobjective} propose to use $q$EHVI with LaMOO, but we opt to use $q$NEHVI instead since it is capable of scaling to the batch size of $q=50$ used in many of our experiments. We refer to this method as LaMOO-$q$NEHVI.
We note that $q$NEHVI is mathematically equivalent to $q$EHVI on noiseless problems. 
The authors propose using rejection sampling to ensure samples come from the ``good'' region.
For high-dimensional search spaces, the acceptance probability is low for uniform random samples from the global design space, and therefore, rejection sampling is prohibitively slow. 
Rejection sampling is used 1) to select starting points for multi-start L-BFGS-B and within the L-BFGS-B routine to enforce that samples are within the ``good'' region. 
We contacted the authors about computational issues with this approach, and the authors recommended to use rejection sampling for selecting starting points, and then to simply run L-BFGS-B from these ``good'' starting points across the global search space. 
With this approach, the resulting candidates may not (and often are not) within the ``good'' region, and LaMOO-qNEHVI is simply an initialization heuristic for optimizing $q$NEHVI, but this approach does speed up candidate generation quite a bit. 
Nevertheless, even using rejection sampling to generate starting points for L-BFGS-B can be (and is on our problems) prohibitively expensive in high-dimensional search spaces. Hence, we limit the rejection sampling by only considering $120,000$ design points before beginning L-BFGS-B with the most promising designs (whether or not they are in the ``good'' region). 
This makes LaMOO-qNEHVI feasible to run our our high-dimensional problems. 

For LaMOO-CMA-ES, we use $q=5$ rather than $q=1$ on vehicle safety, as $q=1$ is not supported.
\subsection{Synthetic problems}
\label{appdx:experimentdetails:synthetic}
The reference points for all problems are given in Table \ref{table:ref_points}. We multiply the objectives (and reference points) for all synthetic problems by $-1$ and maximize the resulting objectives.

\begin{table*}[h]
    \centering
    \begin{small}
    \begin{sc}
    \begin{tabular}{lc}
        \toprule
        Problem & Reference Point\\
        \midrule
        DTLZ2 & [$6$, $6$]\\
        DTLZ3 & $[10^3]^M$\\
        DTLZ5 & $[10]^M$\\
        DTLZ7 & $[15]^M$\\
        Vehicle Safety & [$1698.55$, $11.21$, $0.29$]\\
        Welded Beam & [$40$, $0.015$]\\
        MW7 & [$1.2$, $1.2$] \\
        \bottomrule
    \end{tabular}
    \end{sc}
    \end{small}
    \caption{
    \label{table:ref_points} The reference points for each synthetic benchmark problem.
    }
\end{table*}

\paragraph{DTLZ:} We consider the $2$-objective DTLZ2 problem with various input dimensions $d \in \{10, 30, 100\}$. We also use $2$-objective and $4$-objective variants of DTLZ3, DTLZ5, and DTLZ7 with $d=100$.
The DTLZ problems are standard test problems from the multi-objective optimization literature. 
Mathematical formulas for the objectives in each problem are given in \citet{dtlz}.
\paragraph{MW7:} For a second test problem from the multi-objective optimization literature, we consider a MW7 problem with $2$ objectives, $2$ constraints, and $d=10$ parameters. 
See \citet{mw_test_problems} for details.
\paragraph{Welded Beam:} The welded beam problem \citep{welded_beam} is a structural design problem with $d=4$ input parameters controlling the size of the beam where the goal is to minimize $2$ objectives (cost and end deflection) subject to $4$ constraints. 
More details are given in \citet{tanabe2020}.
\paragraph{Vehicle Safety:} The vehicle safety problem is a $3$-objective problem with $d=5$ parameters controlling the widths of different components of the vehicle's frame. 
The goal is to minimize mass (which is correlated with fuel economy), toe-box intrusion (vehicle damage), and acceleration in a full-frontal collision (passenger injury). 
See \citet{tanabe2020} for additional details.

\subsection{Trajectory planning}
For the trajectory planning, we consider a trajectory specified by $30$ design points that starts at the pre-specified starting location.
Given the $30$ design points, we fit a B-spline with interpolation and integrate over this B-spline to compute the final reward using the same domain as in~\citet{wang2018batched}.
Rather than directly optimizing the locations of the design points, we optimize the difference (step) between two consecutive design points, each one constrained to be in the domain $[0, 0.05] \times [0, 0.05]$.
We use a reference point of [$0$, $0.5$], which means that we want a reward larger than $0$ and a distance that is no more than $0.5$ from the target location [$0.95$, $0.95$].
Since we maximize both objectives, we optimize the distance metric and the corresponding reference point value by $-1$.

\subsection{Optical design}
In order to obtain precise estimates of the optimization performance at reasonable computational cost, we conduct our evaluation on a neural network surrogate model of the optical system rather than on the actual physics simulator. 
The surrogate model was constructed from a dataset of $101$,$000$ optical designs and resulting display images to provide an accurate representation of the real problem. 
The surrogate model is a neural network with a convolutional autoencoder architecture. 
The model was trained using $80$,$000$ training examples and minimizing MSE (averaged over images, pixels, and RGB color channels) on a validation set of $20$,$000$ examples. 
A total of $1$,$000$ examples were held-out for final evaluation. 

\subsection{Mazda vehicle design problem}
We follow the suggestions by \citet{kohira2018proposal} and use the reference point $[1.1, 0]$ and optimize the normalized objectives $\tilde{f}_1 = f_1  - 2$ and $\tilde{f}_2 = f_2 / 74$ corresponding to the total mass and number of common gauge parts, respectively.
Additionally, an initial feasible point is provided with objective values $f_1 = 3.003$ and $f_2 = 35$, corresponding to an initial hypervolume of $\approx 0.046$ for the normalized objectives.
This initial solution is given to all algorithms.
We limit the number of points used for model fitting to only include the $2$,$000$ points closest to the trust region center in case there are more than $2$,$000$ in the larger hypercube with side length $2L$.
Still, for each iteration \ALG using $5$ trust regions fits a total of $56 \times 5$ GP models, a scale far out of reach for any other multi-objective BO method.

\section{Complexity Improvements from Local Modeling}
\label{appdx:complexity_local}
\begin{figure*}[!ht]
    \centering
    \includegraphics[width=0.8\textwidth]{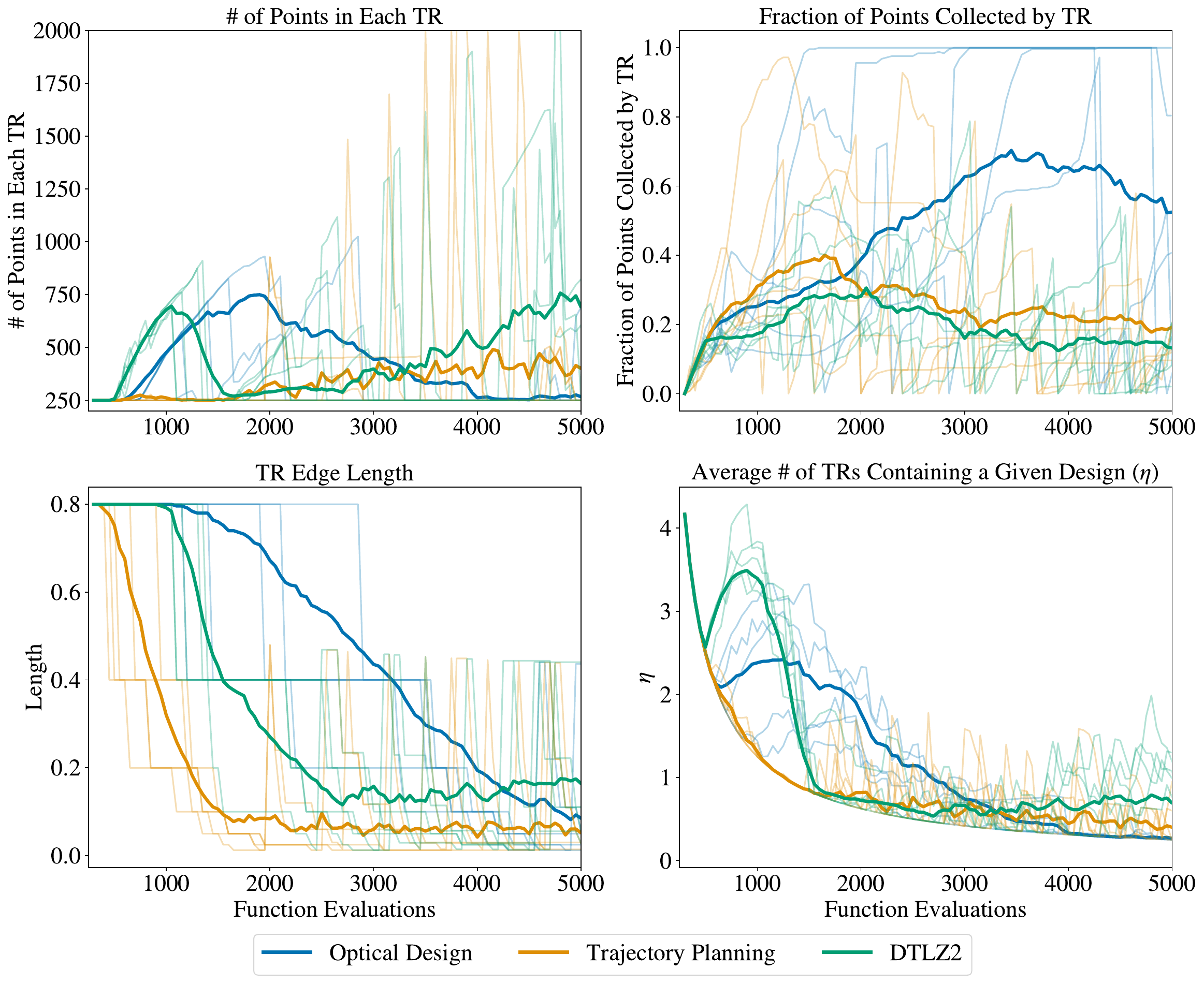}
    \caption{
         For the optical design, trajectory planning, and DTLZ2 problems. We show the average across replications as a solid line and traces from the first replication as transparent lines. (Upper Left) The number of points in each trust region. Trust regions often usually have a few hundred points on average, which results in computationally efficient local modeling. (Upper Right) The number of points in a trust region that was collected by that trust region. This shows that a large fraction of data within a trust region was actually collected by another trust region. (Lower Left) The trust region length. As the optimization proceeds, the trust regions shrink to focus on specific parts of the search space. (Lower Right) The average number of TRs that contain a given design, $\eta \in [0,N_\text{TR}]$. This shows that as the optimization progresses and the TRs shrink, on average less than $1$ TR contains a given design. This is empirical validation of the claim in Appendix~\ref{appdx:complexity_local} that $\eta$ typically becomes small as the optimization progresses and therefore, the complexity improvements are substantial.
    }
    \label{fig:tr_traces}
\end{figure*}

\FloatBarrier
The differences in model fitting time can be even more profound. 
To illustrate this, consider a situation in which a total of $N$ data points have been collected by $n_\text{TR}$ trust regions. Suppose for simplicity that each TR has the same number of observations (under some abuse of nomenclature we use TR to refer to the modeling domain of a TR in this section). Let $\eta$ denote the average number of trust regions that a data point is part of. Then the number of points in each TR is $\eta N/n_\text{TR}$. 
Assuming cubic time complexity for model fitting (i.e. $O(N^3)$ if we used a single global model), the total time complexity of fitting all $n_\text{TR}$ models in the individual TRs is $O\bigl(n_\text{TR} (\eta N/n_\text{TR})^3\bigr) = O\bigl(\eta^3N^3/n_\text{TR}^2\bigr)$. 
This will lead to asymptotic speedups of order $O\bigl(n_\text{TR}^2 / \eta^3\bigr)$ when using local modeling.
Typically, as the optimization progresses and the trust regions shrink, $\eta$ becomes quite small (e.g. $\eta < 1$)\footnote{When $\eta$ is close to the number of trust regions, the ``local" models will fit to nearly all observations, and hence, the models will essentially be global models. The value of $\eta$ at the start of the optimization depends on the initial trust region edge length and the dimension of the search space. }. We validate this claim empirically in the lower right subplot in Figure~\ref{fig:tr_traces}, which shows that $\eta$ becomes less than $1$ on the all problems considered as the optimization progresses. 
In Figure~\ref{fig:tr_traces} we illustrate some additional information from the trust regions to better understand the role of data-sharing and local modeling in \ALG.
Thus, the speedup relative to fitting a single global model can be multiple orders of magnitude.
\subsection{Model fitting times}
\begin{figure}[ht!]
    \centering
    \includegraphics[width=0.4\textwidth]{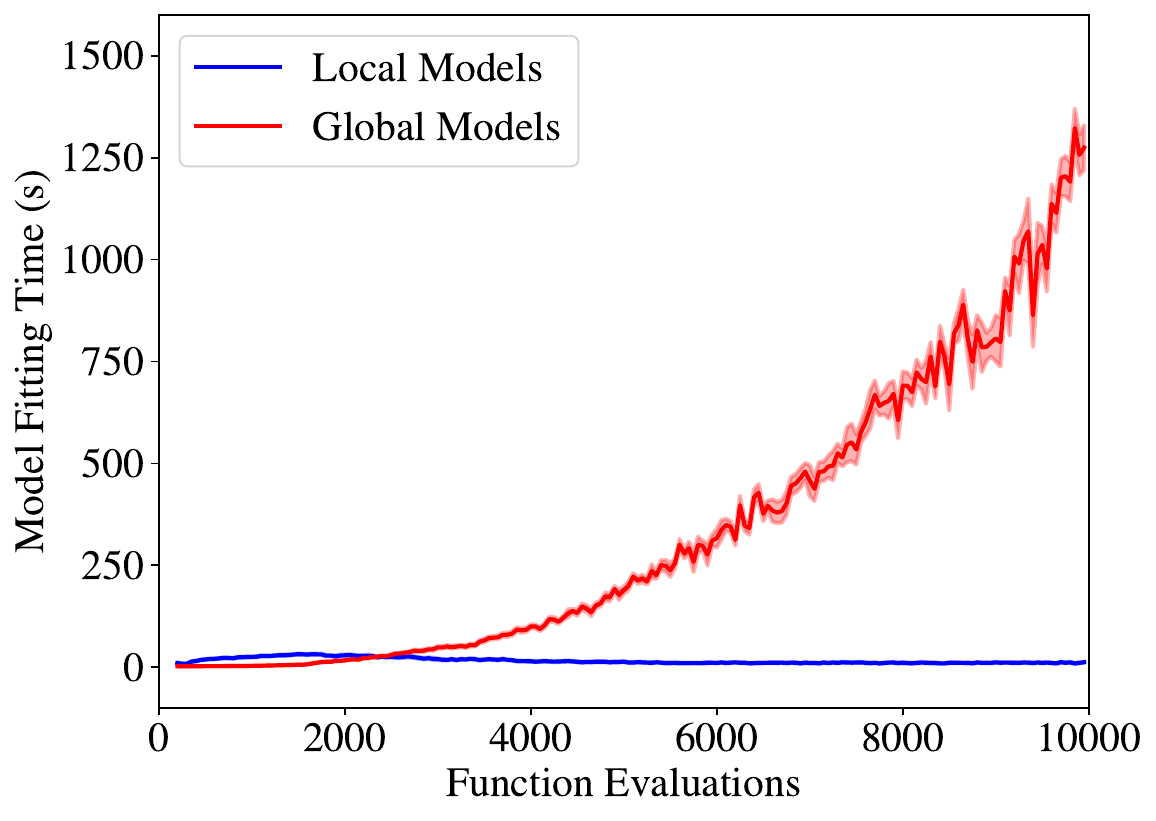}
    \caption{
        Model fitting time for MORBO with local modeling compared to MORBO with one global model on the $146$-dimensional optical design problem. 
        Fitting a global model takes almost~$20$ minutes towards the end of the optimization run compared to~$10$ seconds for \ALG.
        \label{fig:local_vs_global} 
    }
\end{figure}

Empirically, we verify this speedup in Figure~\ref{fig:local_vs_global}.
This can also be seen in the results in Tables~\ref{table:fit_walltimes} and \ref{table:fit_walltimes_dtlz_m4}. 
While candidate generation is fast for TSEMO, the model fitting causes a significant overhead with almost an hour being spent on model fitting after collecting {$2$,$000$} evaluations on the trajectory planning problem. 
This is significantly longer than for \ALG, which only requires far less time for the model fitting due to the use of local modeling.
This shows that the use of local modeling is a crucial component of \ALG that limits the computational overhead from the model fitting. 
The model fitting for \ALG on the optical design problem is less than $25$ seconds while methods such as DGEMO and TSEMO that rely on global modeling require far more time for model fitting after only collecting {$1$,$200$} points.
Additionally, while \ALG needs to fit as many as $56 \times 5 = 280$ GP models on the Mazda problem due to the $54$ black-box constraints and the use of $5$ trust regions, the total time for model fitting still is less than $3$ minutes while this problem is completely out of reach for the other BO methods that rely on global modeling.
\FloatBarrier
\begin{table*}[!ht]
    \centering
    \begin{small}
    \begin{sc} 
    \begin{adjustbox}{max width=\textwidth}
    \begin{tabular}{l|ccc|ccc}
        \toprule
        Problem & DTLZ3 $(M=2)$ & DTLZ5 $(M=2)$ & DTLZ7 $(M=2)$ & DTLZ3 $(M=4)$ & DTLZ5 $(M=4)$ & DTLZ7 $(M=4)$\\
        \midrule
        \ALG{} & 11.0 (0.6) & 9.7 (0.4) & 10.6 (0.4) & 11.5 (0.9) & 10.5 (0.5) & 10.6 (0.4)\\
        NSGA-II & 0.0 (0.0) & 0.0 (0.0) & 0.0 (0.0) & 0.0 (0.0) & 0.0 (0.0) & 0.0 (0.0)\\
        $q$ParEGO & 139.5 (24.6) & 49.1 (2.2) & 26.0 (2.5) & 137.2 (15.4) & 113.2 (6.6) & 49.0 (3.5)\\
        TS-TCH & 64.5 (3.4)& 93.9 (5.8)& 89.6 (3.5) & 143.3 (5.9)& 167.6 (8.8) & 141.3 (6.1) \\
        $q$NEHVI & 133.2 (23.9) & 48.9 (4.9) & 20.8 (1.7) & 25.9 (2.3) & 19.8 (1.7) & 6.8 (0.4)\\
        DGEMO & 5425.1 (142.0) & 1438.0 (29.0) & 180.0 (35.3) & N/A & N/A & N/A \\
        TSEMO & 4246.3 (91.8) & 2481.5 (48.5) & 958.4 (49.1) & 3767.4 (91.0) & 1892.3 (801.5) & 402.0 (31.7)\\
        MOEAD-EGO & 3474.6 (108.6) & 1824.0 (40.1) & 1130.3 (16.0) & 4206.1 (120.5) & 2526.3 (77.5) & 1048.0 (37.8)\\
        \bottomrule
    \end{tabular}
    \end{adjustbox}
    \end{sc}
    \end{small}
    \caption{\label{table:fit_walltimes_dtlz_m4} Model fitting wall time in seconds. The mean and two standard errors of the mean are reported.
    All models were fit on 2x Intel(R) Xeon(R) Gold 6138 CPU @ 2.00GHz. For $M=4$, $q$NEHVI exceeded GPU memory during acquisition optimization and therefore has shorter average model fitting times.}
\end{table*}
\begin{table*}[!ht]
    \centering
    \begin{small}
    \begin{sc}
    \begin{tabular}{l|ccccc}
        \toprule
         Problem & Welded Beam & Vehicle Safety  & Rover& Optical Design & Mazda\\
        \midrule
        \ALG{} & 7.81 (0.02) & 12.58 (0.26) & 9.3 (0.19) & 23.57 (0.36) & 172.53 (1.89)\\
        $q$ParEGO& 0.5 (0.1) & 0.1 (0.0) & 51.6 (16.4) & 46.7 (10.7) & N/A \\
        TS-TCH & 0.5 (0.0) & 0.2 (0.0) & 45.9 (1.8) & 40.5 (4.9) & N/A \\
        $q$NEHVI& 0.5 (0.0) & 0.1 (0.0) & 97.8 (16.3) & 46.4 (3.2) & N/A \\
        DGEMO & N/A & N/A & 809.7 (127.6) & 1109.3 (178.7) & N/A \\
        TSEMO & N/A & 1.0 (0.1) & 305.3 (38.2) & 859.4 (131.4) & N/A \\
        MOEA/D-EGO & N/A & 0.9 (0.0)&373.2 (51.7) & 736.4 (110.4) & N/A \\
        \bottomrule
    \end{tabular}
    \end{sc}
    \end{small}
    \caption{\label{table:fit_walltimes} Model fitting wall time in seconds. The mean and two standard errors of the mean are reported.
    All models were fit on 2x Intel(R) Xeon(R) Gold 6138 CPU @ 2.00GHz. For DGEMO, TSEMO and MOEA/D-EGO only {$1$,$450$} evaluations were performed on Rover (Trajectory Planning) and only {$1$,$250$} evaluations were performed on Optical Design, so the fitting times are shorter than if the full {$2$,$000$} evaluations had been performed.}
\end{table*}
\FloatBarrier

\section{Additional Results}
\label{appdx:additional_results}

\subsection{Low-dimensional problems}
\label{appdx:baby_problems}
\FloatBarrier
We consider two low-dimensional problems to allow for a comparison with existing BO baselines. 
The first problem we consider is a vehicle safety design problem ($d=5$) in which we tune thicknesses of various components of an automobile frame to optimize proxy metrics for maximizing fuel efficiency, minimizing passenger trauma in a full-frontal collision, and maximizing vehicle durability.
The second problem is a welded beam design problem ($d=4$), where the goal is to minimize the cost of the beam and the deflection of the beam under the applied load~\citep{deb2006reference}.
The design variables are the thickness and length of the welds and the height and width of the beam.
In addition, there are $4$ black-box constraints that must be satisfied.

\begin{figure}
    \centering
        \includegraphics[width=0.8\textwidth]{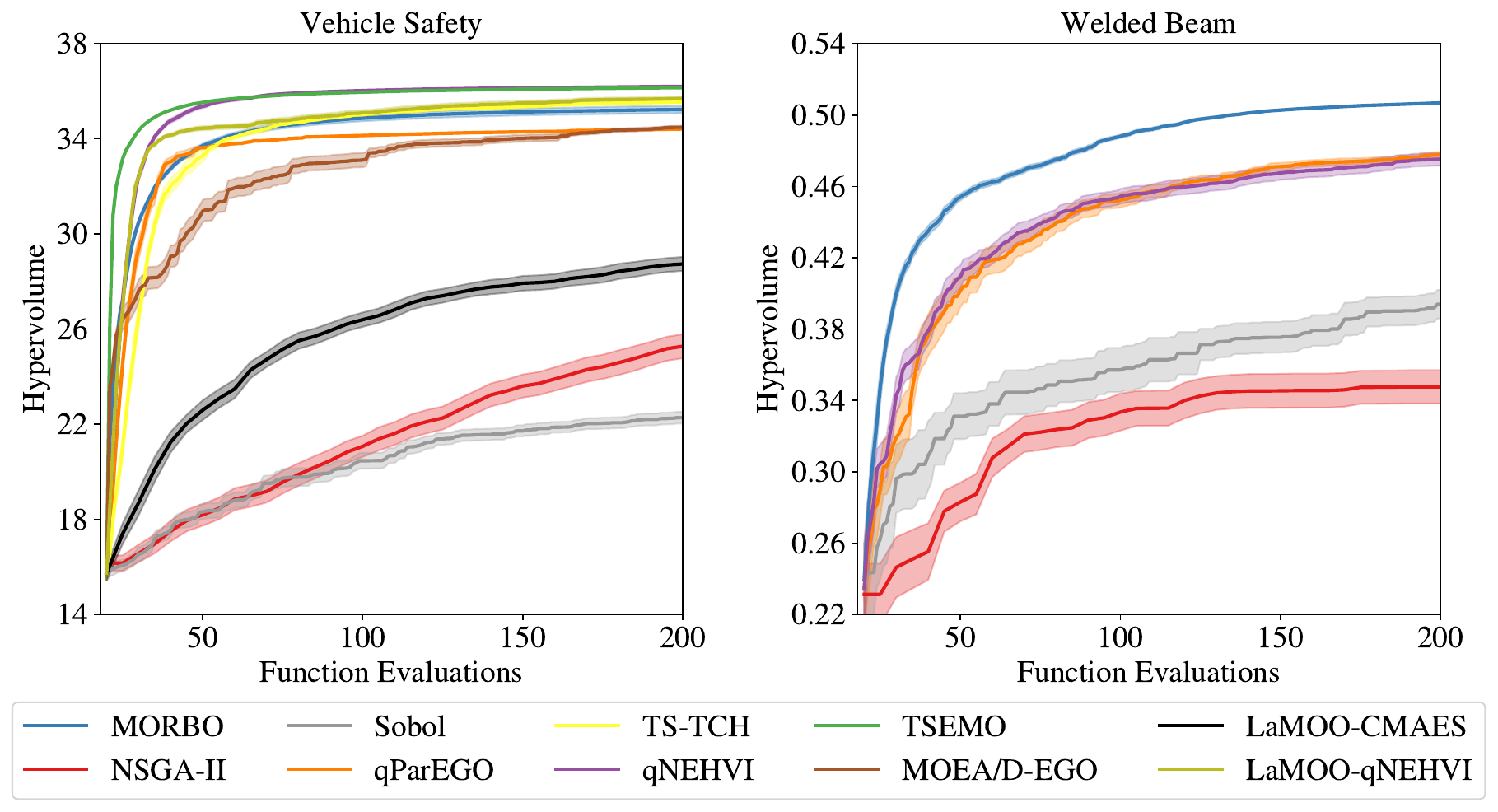}
        \caption{(Left) $q$NEHVI performs the best on the  vehicle design problem ($d=5$) with $3$ objectives. (Right) \ALG outperforms the other methods on  welded beam problem ($d=4$) with $4$ constraints.}
        \label{fig:baby_problems}
\end{figure}

Figure~\ref{fig:baby_problems} presents results for both problems.
While \ALG is not designed for such simple, low-dimensional problems, it is still competitive with other baselines such as TS-TCH and $q$ParEGO on the vehicle design problem, though it cannot quite match the performance of $q$NEHVI and TSEMO.\footnote{DGEMO is not included on this problem as it consistently crashed due to an error deep in the low-level code for the graph-cutting algorithm.}
The results on the welded beam problem illustrate the efficient constraint handling of \ALG.\footnote{DGEMO, TSEMO, MOEA/D-EGO, and TS-TCH are excluded as they do not consider black-box constraints.} 
On both problems, we observe that NSGA-II struggles to keep up, performing barely better (vehicle safety) or even worse (welded beam) than quasi-random Sobol exploration.

\FloatBarrier
\subsection{Candidate Generation Wall Time}
\label{appdx:wall_time_comparisons}
\FloatBarrier
\begin{table*}[!ht]
    \centering
    \begin{small}
    \begin{sc}
    \begin{tabular}{lccccc}
        \toprule
         Problem & Welded Beam & Vehicle Safety  & Rover& Optical Design & Mazda\\
         Batch Size &($q=1$) &($q=1$) & ($q=50$)& ($q=50$) &($q=50$)\\
        \midrule
        \ALG{} & 1.3 (0.0) & 9.6 (0.7) &23.4 (0.4) & 9.8 (0.1)& 188.16 (1.72)\\
        $q$ParEGO &14.5 (0.3)&1.3 (0.0)&213.4 (11.2)&241.9 (14.9)& N/A  \\
        TS-TCH & N/A &0.6 (0.0)&31.3 (1.1)&48.1 (1.2)&N/A \\
        $q$NEHVI &30.4 (0.4)&9.1 (0.1)&997.5 (62.8)&211.27 (6.66)&N/A \\
        NSGA-II &0.0 (0.0)&0.0 (0.0) & 0.0 (0.0) & 0.0 (0.0) & 0.0 (0.0)\\
        DGEMO & N/A & N/A & 697.1 (52.5)&2278.7 (199.8)& N/A \\
        TSEMO & N/A &3.4 (0.1)&3.3 (0.0)& 4.6 (0.1)& N/A \\
        MOEA/D-EGO & N/A & 44.3 (0.3)&71.1 (4.3)&97.5 (6.7)& N/A \\
        LaMOO-CMAES & N/A & 0.6 (0.0) & 2.6 (0.0) & 51.9 (0.3) & N/A \\
        LAMOO-$q$NEHVI& N/A & 24.0 (2.3) & 292.4 (25.2) & 258.8 (1.9) & N/A \\
        \bottomrule
    \end{tabular}
    \end{sc}
    \end{small}
    \caption{\label{table:walltimes} Batch selection wall time (excluding model fitting) in seconds. The mean and two standard errors of the mean are reported.
    \ALG, $q$ParEGO, TS-TCH, and $q$NEHVI were run on a Tesla V100 SXM2 GPU (16GB RAM), while DGEMO, TSEMO, MOEA/D-EGO and NSGA-II were run on 2x Intel(R) Xeon(R) Gold 6138 CPU @ 2.00GHz. For Welded Beam and Vehicle Safety, we ran NSGA-II with $q=5$ in order to avoid a singleton population. For DGEMO, TSEMO and MOEA/D-EGO only {$1$,$450$} evaluations were performed on Rover (Trajectory Planning) and only {$1$,$250$} evaluations were performed on Optical Design, so the generation times are shorter than if the full {$2$,$000$} evaluations had been performed.}
\end{table*}

\begin{table*}[!ht]
    \centering
    \begin{small}
    \begin{sc} 
    \begin{adjustbox}{max width=\textwidth}
    \begin{tabular}{l|ccc|ccc}
        \toprule
        Problem & DTLZ3 $(M=2)$ & DTLZ5 $(M=2)$ & DTLZ7 $(M=2)$ & DTLZ3 $(M=4)$ & DTLZ5 $(M=4)$ & DTLZ7 $(M=4)$\\
        Batch Size &($q=50$) &($q=50$) & ($q=50$) &($q=50$) &($q=50$) & ($q=50$)\\
        \midrule
        \ALG & 26.0 (1.3) & 25.1 (0.9) & 293.0 (21.9) & 976.9 (89.8) & 973.0 (91.8) & 293.0 (21.9)\\
        $q$ParEGO & 315.8 (20.2) & 299.0 (27.2) & 233.0 (21.5) & 372.9 (46.6) & 373.1 (34.6) & 232.4 (22.2)\\
        TS-TCH & 43.6 (1.4) & 49.6 (2.0) & 39.5 (1.9) & 56.5 (1.8) & 69.2 (7.5) & 51.4 (3.4)\\
        $q$NEHVI & 2877.7 (321.3) & 1879.6 (285.4) & 816.9 (49.1) & 4412.9 (600.7) & 3778.2 (266.5) & 57.6 (4.4)\\
        NSGA-II & 0.0 (0.0) & 0.0 (0.0) & 0.0 (0.0) &0.1 (0.0) & 0.0 (0.0) & 0.0 (0.0)\\
        DGEMO & N/A & N/A & N/A & N/A & N/A & N/A \\
        TSEMO & 6.3 (0.1) & 7.2 (0.1) & 6.8 (0.1) & 2878.1 (162.0) & 952.0 (298.1) & 22.2 (3.7)\\
        MOEAD-EGO& 277.8 (1.2) & 224.9 (3.2) & 245.3 (2.9) & 308.7 (2.9) & 303.7 (3.1) & 292.2 (3.5)\\
        \bottomrule
    \end{tabular}
    \end{adjustbox}
    \end{sc}
    \end{small}
    \caption{\label{table:walltimes_dtlz_m4} Batch selection wall time (excluding model fitting) in seconds on DTLZ problems with 2 and 4 objectives with $d=100$. The mean and two standard errors of the mean are reported.}
\end{table*}

While candidate generation time is often a secondary concern in classic BO applications, where evaluating the black box function often takes orders of magnitude longer, existing methods using a single global model and standard acquisition function optimization approaches can become the bottleneck in high-throughput asynchronous evaluation settings that are common with high-dimensional problems.
Tables~\ref{table:walltimes} and \ref{table:walltimes_dtlz_m4} provides a comparison of the wall time for generating a batch of candidates for the different methods on the different benchmark problems.
We observe that the candidate generation for \ALG is two orders of magnitudes faster than for other methods such as $q$ParEGO and $q$NEHVI on the trajectory planning problem where all methods ran for the full {$2$,$000$} evaluations.

\FloatBarrier

\subsection{Pareto Frontiers}
\label{appdx:pareto_frontiers}
\begin{figure}[!ht]
    \centering
    \includegraphics[width=0.88\textwidth]{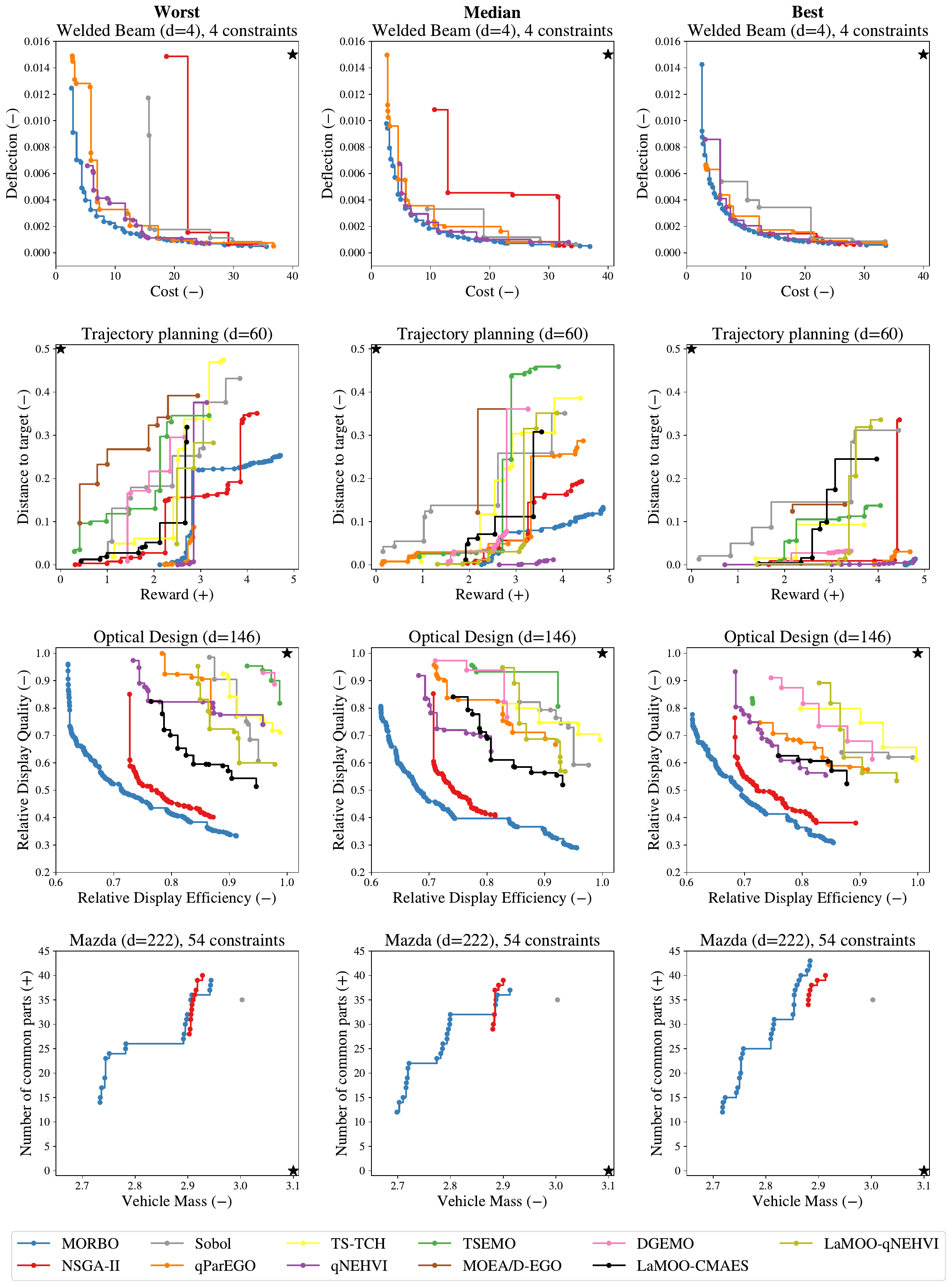}
    \caption{\label{fig:all_pfs}
        In each column we show the Pareto frontiers for the worst, median, and best replications according to the final hypervolume.
        We indicate whether an objective is minimized/maximized by $-/+$, respectively.
        The reference point is illustrated as a black star.
        The use of multiple trust regions allows \ALG to consistently achieve good coverage of the Pareto frontier, in addition to large hypervolumes.
    }
\end{figure}

\FloatBarrier
We show the Pareto frontiers for the welded beam, trajectory planning, optical design, and Mazda problems in Figure~\ref{fig:all_pfs}.
In each column we show the Pareto frontiers corresponding to the worst, median, and best replications according to the final hypervolume.
We exclude the vehicle design problem as it has three objectives which makes the final Pareto frontiers challenging to visualize.

Figure~\ref{fig:all_pfs} shows that even on the low-dimensional $4$D welded beam problem, \ALG is able to achieve much better coverage than the baseline methods.
\ALG also explores the trade-offs better than other methods on the trajectory planning problem, where the best run by \ALG found trajectories with high reward that ended up being close to the final target location.
In particular, other methods struggle to identify trajectories with large rewards while \ALG consistently find trajectories with rewards close to $5$, which is the maximum possible reward.
On both the optical design and Mazda problems, the Pareto frontiers found by \ALG better explore the trade-offs between the objectives compared to NSGA-II and Sobol.
We note that \ALG generally achieves good coverage of the Pareto frontier for both problems.
For the optical design problem, we exclude the partial results found by running the other baselines for $1$k-$2$k evaluations and only show the methods the ran for the full $10$k evaluations.
For the Mazda problem we show the Pareto frontiers of the true objectives and not the normalized objectives that are described in Section~\ref{sec:real_world_problems}.
\ALG is able to significantly decrease the vehicle mass at the cost of using a fewer number of common parts, a trade-off that NSGA-II fails to explore.
It is worth noting that the number of common parts objective is integer-valued and that exploiting this additional information may unlock even better optimization performance of \ALG.

\FloatBarrier
\subsection{Additional Benchmark Problems}
\FloatBarrier
\label{appdx:additional_benchmarks}
To study the performance of \ALG{} on a broader range of problems, we evaluate \ALG{} on two-objective and four-objective versions of DTLZ3, DTLZ5, and DTLZ7 problems with $d=100$. As shown in Figure \ref{fig:dtlz_m4}, \ALG{} performs best on the four-objective DTLZ7 and achieve the best final hypervolume on the four-objective DTLZ3 problem. On the two-objective problems, \ALG{} always ranks in the top 4 methods as shown in Figure \ref{fig:dtlz_m4}. To compare the performance in general across the DTLZ3, DTLZ5, and DTLZ7 problems with a given number of objectives, we rank the methods by the average final hypervolume across replications and compute the average rank across the three problems. As shown in Table \ref{table:rank}, \ALG{} achieves the lowest rank across all methods (which is best) on both M=2 and M=4 problems. DGEMO is not evaluated on the 4-objective problems because the open-source implementation (\url{https://github.com/yunshengtian/DGEMO/tree/master}) does not support more than two objectives. Although DGEMO, MOEA/D-EGO and $q$NEHVI all perform competitively in the two objective setting, all methods are significantly slower than \ALG{}. 

\begin{figure}[!ht]    
    \centering
    \includegraphics[width=0.9\textwidth]{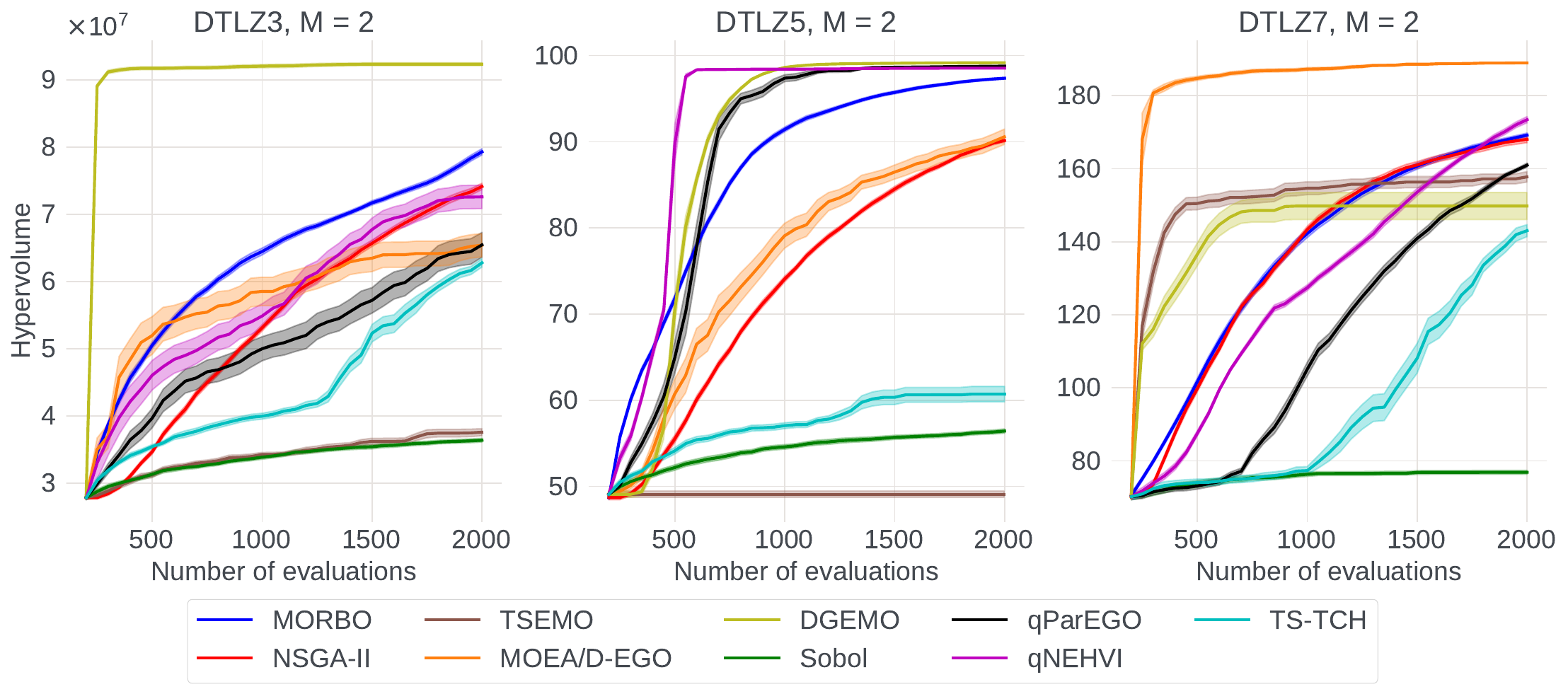}
    \caption{
        \label{fig:dtlz_m2} Optimization performance on two-objective DTLZ3, DTLZ5, and DTLZ7 problems with $d=100$ and $q=50$.
    }
\end{figure}

\begin{figure}[!ht]
    \centering
    \includegraphics[width=0.9\textwidth]{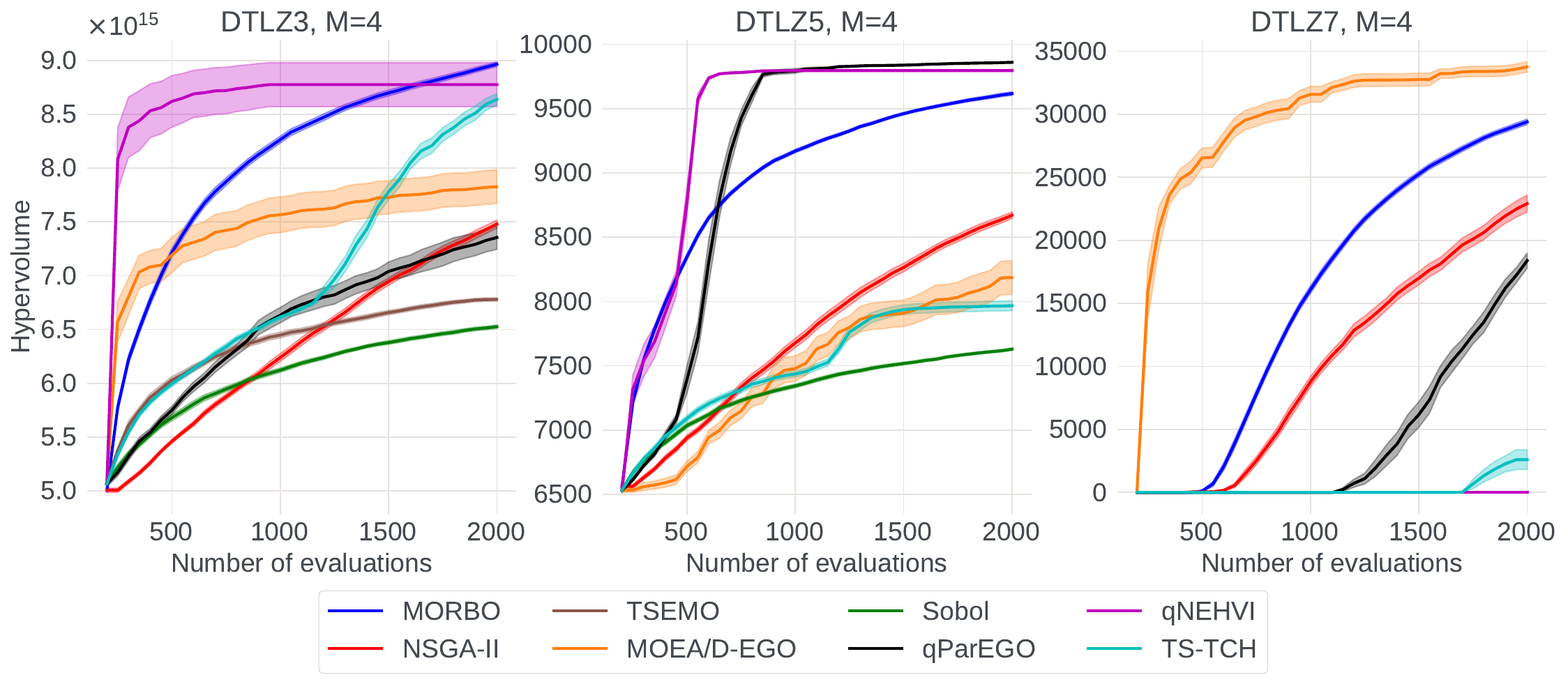}
    \caption{
        \label{fig:dtlz_m4} Optimization performance on four-objective DTLZ3, DTLZ5, and DTLZ7 problems with $d=100$ and $q=50$.
    }
\end{figure}

\begin{table*}[!ht]
    \centering
    \begin{small}
    \begin{sc}
    \begin{small}
    \begin{tabular}{lcc}
        \toprule
         & Avg. Rank for M=2 & Avg. Rank for M=4 \\
        \midrule
        \ALG{} & 3.0 & 1.67\\
        $q$ParEGO & 4.0 & 3.3  \\
        $q$NEHVI & 3.0 & 3.16  \\
        TS-TCH & 7.3 & 4.3 \\
        NSGA-II & 4.3 & 3.7\\
        DGEMO & 3.0 & 8.2 \\
        TSEMO & 7.7 & 8.3 \\
        MOEA/D-EGO & 4.0 & 5.5 \\
        Sobol & 8.7 & 6.8 \\
        \bottomrule
    \end{tabular}
    \end{small}
    \end{sc}
    \end{small}
    \caption{\label{table:rank} Mean rank across DTLZ3, DTLZ5, and DTLZ7 problems based on final mean hypervolume with $d=100$ and $q=50$. A lower rank means the method achieves better final performance on average across the DTLZ3, DTLZ5, and DTLZ7 problems with $M$ objectives.}
\end{table*}

\end{document}